\documentclass[acmsmall]{acmart}
\microtypesetup{expansion=false}
\renewcommand\footnotetextcopyrightpermission[1]{}
\usepackage{tikz}
\usetikzlibrary{arrows.meta,positioning,fit,calc,shapes.geometric,matrix,backgrounds}
\usepackage{tabularx}
\usepackage{xcolor}
\usepackage{tikz,forest,graphicx}
\usetikzlibrary{arrows.meta,positioning,fit,calc}
\useforestlibrary{edges}

\title{Towards Reliable AI Data Scientists: Data Agents with Workflow Harnesses}
\begin{document}

\author{Huachi Zhou}
\affiliation{%
  \institution{The Hong Kong Polytechnic University}
  \city{Hong Kong}
  \country{China}
}
\email{huachi.zhou@connect.polyu.hk}

\author{Yujing Zhang}
\affiliation{%
  \institution{The Hong Kong Polytechnic University}
  \city{Hong Kong}
  \country{China}
}
\email{yu-jing.zhang@connect.polyu.hk}

\author{Jiahe Du}
\affiliation{%
  \institution{The Hong Kong Polytechnic University}
  \city{Hong Kong}
  \country{China}
}
\email{jiahe.du@connect.polyu.hk}

\author{Jiacheng Cai}
\affiliation{%
  \institution{The Hong Kong Polytechnic University}
  \city{Hong Kong}
  \country{China}
}
\email{jiacheng.cai@connect.polyu.hk}

\author{Zijin Hong}
\affiliation{%
  \institution{The Hong Kong Polytechnic University}
  \city{Hong Kong}
  \country{China}
}
\email{zijin.hong@connect.polyu.hk}

\author{Chuang Zhou}
\authornote{Corresponding author.}
\affiliation{%
  \institution{The Hong Kong Polytechnic University}
  \city{Hong Kong}
  \country{China}
}
\email{chuang-qqzj.zhou@connect.polyu.hk}

\author{Zheng Yuan}
\authornotemark[1]
\affiliation{%
  \institution{The Hong Kong Polytechnic University}
  \city{Hong Kong}
  \country{China}
}
\email{yzheng.yuan@connect.polyu.hk}

\author{Qinggang Zhang}
\affiliation{%
  \institution{Jilin University}
  \city{Changchun}
  \country{China}
}
\email{qinggangzhang@jlu.edu.cn}

\author{Qing Li}
\affiliation{%
  \institution{The Hong Kong Polytechnic University}
  \city{Hong Kong}
  \country{China}
}
\email{qing-prof.li@polyu.edu.hk}

\author{Xiao Huang}
\affiliation{%
  \institution{The Hong Kong Polytechnic University}
  \city{Hong Kong}
  \country{China}
}
\email{xiao.huang@polyu.edu.hk}

\authorsaddresses{%
Huachi Zhou, Yujing Zhang, Jiahe Du, Jiacheng Cai, Zijin Hong,
Chuang Zhou, Zheng Yuan, Qing Li, and Xiao Huang,
The Hong Kong Polytechnic University, Hong Kong, China.
Emails: \{huachi.zhou, yu-jing.zhang, jiahe.du, jiacheng.cai,
zijin.hong, chuang-qqzj.zhou, yzheng.yuan\}@connect.polyu.hk;
\{qing-prof.li, xiao.huang\}@polyu.edu.hk.
Qinggang Zhang,
Jilin University, Changchun, China.
Email: qinggangzhang@jlu.edu.cn.
}

\renewcommand{\shortauthors}{Zhou, et al.}

\begin{abstract}
Large language model agents are increasingly deployed for data-intensive work, yet reliable data analysis requires more than general-purpose reasoning and ad hoc tool augmentation. Data Agents, equipped with workflow harnesses, offer a promising paradigm for automating the end-to-end data science lifecycle. This survey reviews Data Agents from a harness-centric perspective. First, we introduce a taxonomy of Data Agents and associated data environments, organizing the literature around five functional stages: perception, planning, execution, verification, and repair. Second, we analyze the key technical routes within each stage, identifying 15 distinct approaches ranging from data structure probing to data state reconstruction. Third, we identify four open reliability problems: inactive semantic calibration, missing clarification, missing experience transfer, and the missing verification-repair repository. These problems explain why silent failures can persist even when individual components function correctly, highlighting the need for rigorous workflow harnesses and shared reliability resources. Finally, we summarize the horizontal task families of Data Agents, examine their vertical application settings, and review benchmarks for evaluation, while maintaining a companion repository at https://github.com/DEEP-PolyU/Awesome-Data-Agents
\end{abstract}

\maketitle

\section{Introduction}

Large language model (LLM) agents have rapidly emerged as a primary interface for data-intensive work~\cite{zhu2026surveydataagentsemerging,fu2025autonomousdataagentsnew,zhou2025surveyllmtimesdata}. Their strength lies in general-purpose reasoning applied to data science tasks: LLMs excel at analyzing tasks, generating plausible plans, translating natural language into executable data operations such as SQL or Python, and adapting to diverse tasks with minimal examples. This capability makes LLM agents a natural bridge between data science tasks and complex data environments~\cite{Sun2025surveylargelanguage_106,Luo2025naturallanguagestate_109}.

However, general-purpose reasoning does not inherently guarantee analytical reliability. Reliability requires logically correct outputs in real-world data science tasks, even in imperfect data environments. These environments introduce five recurring challenges that hinder LLM-based agents, even when the underlying LLM possesses strong reasoning capabilities:

\textbf{First, true data semantics are often inadequately represented in surface-level contexts.} Data is typically exposed to agents through raw formats, spreadsheet layouts, or feature definitions in ML pipelines, which provide only partial representations of its underlying semantics. Important information, such as semantic relationships among fields and governing properties such as data-generating distributions, may therefore remain poorly encoded for downstream reasoning~\cite{lin2026semasqltraditionalrelationalquerying,Peeters2025entitymatchingusing_133}. 

\textbf{Second, the solution space is massively underdetermined.} For a given task and data environment, there is rarely a single canonical analytical solution. Instead, the task can be addressed through countless strategies, making it difficult to identify the optimal solution~\cite{hu2025mctsrag,lu2026octotools,otani2026agents}.

\begin{figure*}[t]
    \centering
    \includegraphics[width=\linewidth]{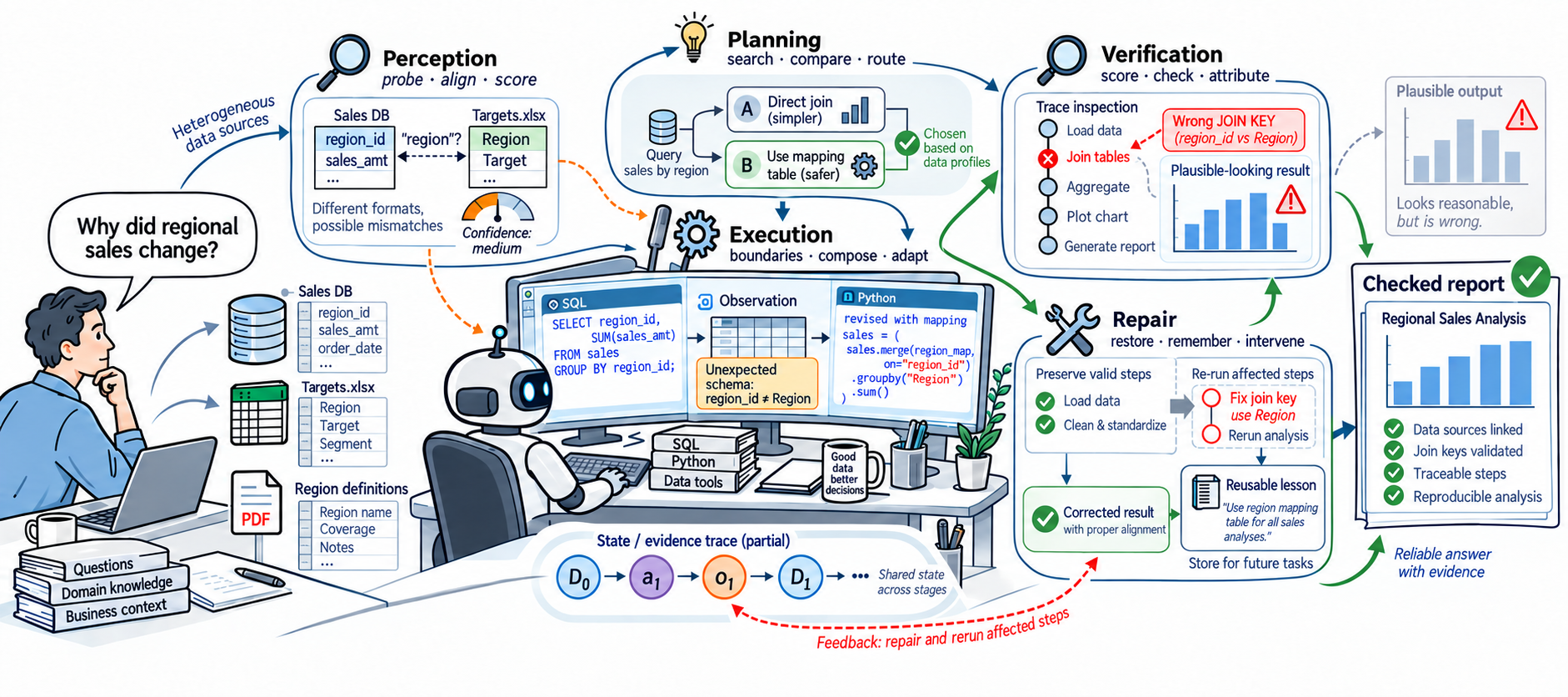}
    \caption{The workflow harness of AI Data Scientists. The harness governs Data Agent workflows through five functional stages, including perception, planning, execution, verification, and repair. These stages are invoked iteratively as needed to maintain the evolving data state across data tasks.}
    \label{fig:Workflow}
\end{figure*}

\textbf{Third, effective tool execution is inherently dynamic.} The optimal next step depends on intermediate observations that only become available after a previous tool is executed. Dynamic tool inputs and outputs make the optimal next step a moving target, preventing the full execution sequence from being determined in advance~\cite{chen2025learning,huang2026toolomnienablingopenworld,lu2026octotools}.

\textbf{Fourth, failure signals are sparse and non-local.} A data science task can complete successfully and produce a plausible-looking result without raising any exception, even when the result is grounded in flawed assumptions. Conversely, when execution does fail explicitly, the observed error often reflects only a downstream symptom, providing little evidence about which earlier step introduced the underlying problem~\cite{rewolinski2026sanitychecksagenticdata,lin2026reflectinterventionsupportederrorattribution}.

\textbf{Fifth, repairs are entangled with local data context.} A code repair is typically tied to the specific data context, e.g., the schema and spreadsheet layout for analytical tasks, or the value distributions that determine appropriate algorithms for modeling tasks. The underlying logical correction is deeply entangled with this local context, making it difficult to reuse the repair code~\cite{zhuang2026agentrewind,lin2026reflectinterventionsupportederrorattribution,su2026failure}.

These five challenges motivate the need for a disciplined approach beyond simply equipping LLMs with more tools or stronger reasoning capabilities. Prior advances have often treated the agent as a collection of capabilities rather than as a system that follows a disciplined analytical process. Yet, an agent with perfect tools and reasoning can still produce wrong answers silently, not because it lacks ability, but because it lacks a workflow to guide its behavior. Without such a workflow, the agent remains a brittle black box, particularly in data environments where the most common failures produce no explicit error.

Data Agents have emerged as an autonomous response to these difficulties~\cite{li2026deepeyesteerableselfdrivingdata,roy2026cedarcontextengineeringagentic}. To succeed, they rely on what we term a workflow harness: not a rigid pipeline, but a disciplined, iterative framework that governs the workflow through five functional stages, including perception, planning, execution, verification, and repair. Within this harness, the agent exercises judgment by invoking the appropriate functional stages as needed, transforming from a one-shot answer generator into a reliable AI Data Scientist. To operate effectively within this harness, a Data Agent pursues five interconnected goals:

\textbf{Deep Semantic Grounding.}
To capture the true semantics underlying surface-level data, Data Agents infer the intrinsic structures, e.g., semantic relationships, and governing properties, e.g., distributional properties from heterogeneous sources~\cite{shahbazi2026omnitqacostawarehybridquery,aminnaseri2026bluedataintelligencelayer}. Since the full data volume may exceed input limits, agents transform these raw inputs into a semantically grounded representation that enables correct downstream reasoning.

\textbf{Efficient Solution Space Navigation.}
To navigate the massively underdetermined solution space, Data Agents decompose high-level tasks into executable steps, evaluate alternative steps, and plan under uncertainty. They efficiently prune the action space and accelerate convergence toward near-optimal solutions within practical time constraints~\cite{yang2026tooltreeefficientllmagent,parekh2026pexaparallelexplorationagent}.

\textbf{Dynamic Tool Orchestration.}
To handle the dynamic nature of tool execution, Data Agents understand what tools are available, dynamically compose them into coherent orchestration sequences, and adapt their tool-use strategy on the fly as intermediate observations arrive~\cite{lu2026octotools,chen2025learning}.

\textbf{Routine Verifiability.}
Silent failures often remain undetected until the final result emerges, by which point the error may have already propagated through multiple preceding steps. Data Agents need routine checking that allows timely inspection of key analytical steps, so that errors can be caught at their source before they cascade~\cite{lin2026reflectinterventionsupportederrorattribution,rewolinski2026sanitychecksagenticdata}.

\textbf{Generalized Repair.}
To overcome context-bound repairs, Data Agents must disentangle the underlying logical correction from the local context and abstract the repair into a generalized lesson, e.g., a reusable code template~\cite{zhuang2026agentrewind,zhang2025deepanalyzeagenticlargelanguage}. This distills the correction into reusable knowledge, so that lessons learned from one task can transfer across structurally or statistically different data. 

Recent work has begun to move in this direction: the literature is shifting from isolated query answering toward agentic data analysis over heterogeneous sources, and from static prompting toward feedback-driven iterative refinement~\cite{wang2026fdabenchbenchmarkdataagents,ma2026aiagentsanswerdata,li2026deepeyesteerableselfdrivingdata}. Yet these advances remain highly fragmented. Most existing works focus on a narrow modality, task, or stage of the analytical workflow, with limited integration across the full data analysis process. As a result, they provide only partial support for cross-modal or long-horizon analyses, where different stages must remain coordinated over time. This fragmentation also obscures the common reliability challenges shared across otherwise disparate systems. It therefore motivates a systematic synthesis of the literature through the lens of a unified workflow harness, which is the focus of this survey.
\begin{itemize}
  \item First, we establish a formal foundation for Data Agents by defining the heterogeneous data environment and modeling the agent's operational trajectory. Then we conceptualize the necessity of a disciplined workflow harness to achieve reliability in data science tasks and mitigate silent failures, distinguishing our work from existing literature (Section 2).
  \item Second, we introduce a five-stage workflow harness consisting of perception, planning, execution, verification, and repair, and systematically classify the technical methods within each stage into 15 distinct routes. Each stage is defined by a functional objective, and its three technical routes capture complementary mechanisms through which existing Data Agents realize that objective (Section~3).
    \item Third, we identify four reliability problems that reveal why existing harnesses remain insufficient, and what future harness designs must address to overcome them (Section 4).
    \item Fourth, we complement the workflow-harness analysis by examining what analytical capabilities Data Agents perform (Section 5), how these capabilities are composed in vertical application settings (Section 6), and how both capabilities and workflow reliability are evaluated by existing benchmarks (Section 7).
\end{itemize}

\section{Background and Problem Formulation for Data Agents}

\subsection{Evolution of the AI Data Scientist Paradigm}

The emergence of AI Data Scientists represents a paradigm shift in how we automate data science. Earlier generations of automation, such as AutoEDA systems, streamlined basic data profiling and visualization; however, their outputs were typically static reports rather than dynamic, reasoning-driven analytical processes. Subsequently, AutoML frameworks automated model selection, hyperparameter tuning, and architecture search, yet they predominantly operated under the rigid assumption that input data had already been cleaned and transformed. More recent AutoResearch systems advanced beyond mere model selection by focusing on iterative training code modification, though they generally lack a comprehensive workflow design for end-to-end data tasks. Concurrently, code agents represented a further step by translating natural language instructions into executable code, often operating within a basic harness that supports execution, testing, and repair. While these preceding systems laid a useful foundation, they remain insufficient for solving data science tasks reliably. Building on this evolution, the current paradigm of Data Agents emerges to bridge these gaps, integrating advanced reasoning with a disciplined workflow harness to navigate real-world, heterogeneous data environments.

\subsection{Formal Definition of the Data Environment}

Before introducing the workflow, we first formally define the environment where Data Agents operate. We model the data environment as a finite collection of $m$ heterogeneous data sources: $$
\mathcal{D} = \{d_1, d_2, \ldots, d_m\}
$$

Each data source \( d_j \) extends beyond a simple text stream, manifesting as a structured tuple: $$
d_j = \langle X_j, \tau_j, \Sigma_j, \Pi_j \rangle
$$
where \( X_j \) denotes the \textit{data payload} (i.e., the raw content, such as cell values in a CSV, rows in a database, pixels in a chart, or text in a document); \( \tau_j \) indicates the \textit{structural type} (structured, semi-structured, or unstructured); \( \Sigma_j \) captures the \textit{data structure} (e.g., column schemas for relational data, nested key paths for JSON, or layout and header positions for spreadsheets); and \( \Pi_j \) represents the \textit{data properties} (e.g., unique constraints, formula dependencies, or statistical distributions characterizing the values). Together, \( \Sigma_j \) and \( \Pi_j \) define the \textit{data context} that the agent must navigate. The collective configuration of all data sources, encompassing their payloads, structures, and properties, constitutes the current \textit{data state} of the environment.

\paragraph{Structured Data.}
For structured sources, such as relational databases and time-series data, \( \Sigma_j \) is explicit. \( \Pi_j \) is well-defined through integrity constraints and domain rules. A Data Agent can query such data through deterministic operations, e.g., SQL for databases or time-series queries for temporal data. Research problems include aligning \( \Sigma_j \) with task intent.

\paragraph{Semi-structured Data.}
For semi-structured sources, e.g., spreadsheets, HTML, and JSON, \( \Sigma_j \) is partial: headers may be missing, and nesting may be arbitrary, while \( \Pi_j \) is accessible, such as formula dependencies in spreadsheets or the structural logic of JSON. A Data Agent first infers \( \Sigma_j \), then handles structural variability in later stages. Research problems include reliable structure inference and handling of hybrid \( \Pi_j \) such as cross-sheet references.

\paragraph{Unstructured Data.}
For unstructured sources, such as PDFs, images, and raw text, \( \Sigma_j \) is absent, \( \Pi_j \) is latent, and the payload \( X_j \) is the only source of information. A Data Agent performs active extraction, constructing task-dependent structured representations from free text, document layouts, or chart objects. There is no universal rule set \( \Pi_j \) for unstructured content. Research efforts focus on cross-modal grounding and layout-aware parsing to derive \( \Sigma_j \) and \( \Pi_j \) for reasoning.

\definecolor{colorPerception}{HTML}{4A9DAF}
\definecolor{colorPlanning}{HTML}{D6A84B}
\definecolor{colorExecution}{HTML}{78A65A}
\definecolor{colorVerification}{HTML}{A28CC2}
\definecolor{colorRepair}{HTML}{D77A82}

\tikzset{
    taxonomy-box/.style={rectangle, rounded corners=1.5pt, align=left,
        text=black, font=\scriptsize, inner xsep=2pt, inner ysep=2pt,
        line width=0pt},
    taxonomy-root/.style={taxonomy-box, draw=gray, fill=gray!12,
        font=\scriptsize\bfseries},
    stage-perception/.style={taxonomy-box, fill=colorPerception!28, font=\scriptsize\bfseries},
    stage-planning/.style={taxonomy-box, fill=colorPlanning!28, font=\scriptsize\bfseries},
    stage-execution/.style={taxonomy-box, fill=colorExecution!28, font=\scriptsize\bfseries},
    stage-verification/.style={taxonomy-box, fill=colorVerification!28, font=\scriptsize\bfseries},
    stage-repair/.style={taxonomy-box, fill=colorRepair!28, font=\scriptsize\bfseries},
    type-perception/.style={taxonomy-box, fill=colorPerception!15, text width=8.4em},
    type-planning/.style={taxonomy-box, fill=colorPlanning!15, text width=8.4em},
    type-execution/.style={taxonomy-box, fill=colorExecution!15, text width=8.4em},
    type-verification/.style={taxonomy-box, fill=colorVerification!15, text width=8.4em},
    type-repair/.style={taxonomy-box, fill=colorRepair!15, text width=8.4em},
    works-perception/.style={taxonomy-box, fill=white, text width=26em},
    works-planning/.style={taxonomy-box, fill=white, text width=26em},
    works-execution/.style={taxonomy-box, fill=white, text width=26em},
    works-verification/.style={taxonomy-box, fill=white, text width=26em},
    works-repair/.style={taxonomy-box, fill=white, text width=26em}
}

\forestset{
    works-perception/.style={draw=colorPerception, fill=white, line width=0.4pt,
        rounded corners=1.5pt, text width=26em, align=left,
        font=\scriptsize, inner xsep=2pt, inner ysep=2pt},
    works-planning/.style={draw=colorPlanning, fill=white, line width=0.4pt,
        rounded corners=1.5pt, text width=26em, align=left,
        font=\scriptsize, inner xsep=2pt, inner ysep=2pt},
    works-execution/.style={draw=colorExecution, fill=white, line width=0.4pt,
        rounded corners=1.5pt, text width=26em, align=left,
        font=\scriptsize, inner xsep=2pt, inner ysep=2pt},
    works-verification/.style={draw=colorVerification, fill=white, line width=0.4pt,
        rounded corners=1.5pt, text width=26em, align=left,
        font=\scriptsize, inner xsep=2pt, inner ysep=2pt},
    works-repair/.style={draw=colorRepair, fill=white, line width=0.4pt,
        rounded corners=1.5pt, text width=26em, align=left,
        font=\scriptsize, inner xsep=2pt, inner ysep=2pt}
}

\begin{figure*}[t]
    \centering
    \resizebox{\textwidth}{!}{%
        \begin{forest}
            forked edges,
            for tree={
                grow=east,
                reversed=true,
                anchor=base west,
                parent anchor=east,
                child anchor=west,
                base=left,
                rectangle,
                draw=none,
                rounded corners,
                align=left,
                edge+={darkgray!45, line width=0.6pt},
                s sep=2.5pt,
                l sep=7pt,
                inner xsep=1pt,
                inner ysep=1.5pt,
                ver/.style={rotate=90, child anchor=north, parent anchor=south, anchor=center}
            },
            [{Data Agents}, taxonomy-root, ver
                [{Perception}, stage-perception, text width=6.6em
                    [{Structured Data}, type-perception
                        [{Arming Data Agents~\cite{agarwal2026arming}; PV-SQL~\cite{tian2026pvsqlsynergizingdatabaseprobing}; FlexSQL~\cite{pham2026flexsql}; \\ TSQAgent~\cite{wu2026tsqagentratingtimeseries}; TimeClaw~\cite{liu2026timeclawtimeseriesaiagent};\\ OmniTQA~\cite{shahbazi2026omnitqacostawarehybridquery}; Cortex AISQL~\cite{liskowski2026cortex}; \\ SEMA-SQL~\cite{lin2026semasqltraditionalrelationalquerying}; Blue DIL~\cite{aminnaseri2026bluedataintelligencelayer};\\ LLM Entity Matching~\cite{Peeters2025entitymatchingusing_133}; Agent-OM~\cite{Qiang2024agentleveragingagents_84}; $R^3$-NL2GQL~\cite{Zhou2024nl2gqlmodelcoordination_37}}, works-perception]
                    ]
                    [{Semi-Structured Data}, type-perception
                        [{TabClaw~\cite{cheng2026tabclawinteractiveselfevolvingagent}; TableMind~\cite{jiang2026tablemind}; Cocoon~\cite{Huang2024cocoonsemantictable_122}; \\ RubikSQL~\cite{chen2025rubiksqllifelonglearningagentic}; AgenticData~\cite{sun2025agenticdata};\\ TwinBI~\cite{li2026twinbiagenticdigitaltwin}; STA Agent~\cite{geng2025llmagentbasedcomplexsemantic}; TABQAWORLD~\cite{kwok2026tabqaworldoptimizingmultimodalreasoning}; \\ SemPiper~\cite{ovcharenko2026sempiperinteractivecodesynthesis}; DataSTORM~\cite{liu2026datastormdeepresearchlargescale};\\ Beyond Linear LLM Invocation~\cite{hou2026linearllminvocationefficient}; CoDA~\cite{chen2025codaagenticsystemscollaborative}}, works-perception]
                    ]
                    [{Unstructured Data}, type-perception
                        [{MimirRAG~\cite{samuelsen2026mimirragmultiagentragframework}; MAVEN~\cite{zhang2026mavenmultistageagenticannotation}; \\ HARMON-E~\cite{gupta2025harmonehierarchicalagenticreasoning}; Navigating the Mirage~\cite{zhang2026navigatingmiragedualpathagentic};\\ FinAgent-RAG~\cite{shu2026agenticretrievalaugmentedgenerationfinancial}; \\ Code-in-the-Loop Forensics~\cite{zhang2026codeintheloopforensicsagentictool}; LongDS-Bench~\cite{xu2026longdsbenchfailurelonghorizonagentic};\\ Doc-Researcher~\cite{dong2025docresearcherunifiedmultimodaldocument}}, works-perception]
                    ]
                ]
                [{Planning}, stage-planning, text width=6.6em
                    [{Structured Data}, type-planning
                        [{MARS-SQL~\cite{yang2026marssqlmultiagentreinforcementlearning}; FlexSQL~\cite{pham2026flexsql}; CHASE-SQL~\cite{Pourreza2025chasemultipath_83}; \\ OpenSearch-SQL~\cite{Xie2025opensearchenhancingtext_215}; EllieSQL~\cite{zhu2025elliesql};\\ SiriusBI~\cite{Jiang2025siriusbicomprehensivepowered_159}; PExA~\cite{parekh2026pexaparallelexplorationagent}; Insight Agents~\cite{bai2025insight}}, works-planning]
                    ]
                    [{Semi-Structured Data}, type-planning
                        [{Jupiter~\cite{li2026jupiter}; SemPipes~\cite{ovcharenko2026sempipesoptimizablesemantic}; ArchPilot~\cite{yuan2025archpilotproxyguidedmultiagentapproach}; \\ Mixture-of-Minds~\cite{zhou2025mixture}; TaTToo~\cite{zou2026tattoo};\\ DataPRM~\cite{qiu2026rewarding}; ELLM-FT~\cite{Gong2025evolutionarylargelanguage_44}; \\ RetClean~\cite{Naeem2024retcleanretrievalbased_48}; Dataforge~\cite{wang2026dataforgeagenticplatformautonomous}; Chat2Data~\cite{Zhao2024chat2datainteractivedata_225}}, works-planning]
                    ]
                    [{Unstructured Data}, type-planning
                        [{MCTS-RAG~\cite{hu2025mctsrag}; DocETL~\cite{Shankar2025docetlagenticquery_198}; UQE~\cite{Dai2024queryengineunstructured_164}; \\ FH vs. SH~\cite{otani2026agents}; FinAgent-RAG~\cite{shu2026agenticretrievalaugmentedgenerationfinancial};\\ DeepXiv-SDK~\cite{qian2026deepxivsdkagenticdatainterface}; Doc-Researcher~\cite{dong2025docresearcherunifiedmultimodaldocument}; AgenticScholar~\cite{lan2026agenticscholar}}, works-planning]
                    ]
                ]
                [{Execution}, stage-execution, text width=6.6em
                    [{Tool-Requirement Signals}, type-execution
                        [{AlloBench~\cite{wang2026allobenchmeasuringonline}; VisualToolBench~\cite{guo2025seeingevaluatingmultimodalllms}; TimeART~\cite{wu2026timeartagentictimeseries}; Cast-R1~\cite{tao2026castr1learningtoolaugmentedsequential}; TimeClaw~\cite{liu2026timeclawtimeseriesaiagent}}, works-execution]
                    ]
                    [{Tool-Specification Signals}, type-execution
                        [{EasyTool~\cite{yuan2025easytool}; NaviAgent~\cite{jiang2026naviagent}; \\ OctoTools~\cite{lu2026octotools}; ToolEVO~\cite{chen2025learning}; DeepEye~\cite{li2026deepeyesteerableselfdrivingdata};\\ Analytic Agent~\cite{singh2026texttosqlagenticllmgoverned}}, works-execution]
                    ]
                    [{Tool-Feedback Signals}, type-execution
                        [{ToolFuzz~\cite{milev2025toolfuzzautomatedagent}; ToolOmni~\cite{huang2026toolomnienablingopenworld}; \\ OpenAgent~\cite{lv2026agentsgeneralizeopenworld}; Structured Reflection~\cite{su2026failure};\\ ToolCritic~\cite{hamad2025toolcriticdetectingcorrecting}; \\ On-Policy Data Evolution~\cite{huang2026onpolicydataevolutionvisualnative}; ReTools~\cite{dong2026retoolsreflectionenhancedtool}; SQL-Trail~\cite{hua2026sqltrailmultiturnreinforcementlearning}}, works-execution]
                    ]
                ]
                [{Verification}, stage-verification, text width=6.6em
                    [{Structured Data}, type-verification
                        [{MERIT~\cite{wang2026learningretrieveduallevellongterm}; Reward-SQL~\cite{zhang2026rewardsqlboostingtexttosql}; CHASE-SQL~\cite{Pourreza2025chasemultipath_83}; OpenSearch-SQL~\cite{Xie2025opensearchenhancingtext_215}; \\ E2ETune~\cite{Huang2025e2etuneknobtuning_112};
                         PMVisAgent~\cite{xu2026reliableagenticprogressivetexttovisualization};\\ $\lambda$-Tune~\cite{Giannakouris2025tuneharnessinglarge_111}; CausalFlow~\cite{bonagiri2026causalflowcausalattributioncounterfactual}; REFLECT~\cite{lin2026reflectinterventionsupportederrorattribution}; PExA~\cite{parekh2026pexaparallelexplorationagent}}, works-verification]
                    ]
                    [{Semi-Structured Data}, type-verification
                        [{Table-Critic~\cite{Yu2025tablecriticmulti_151}; TabTracer~\cite{luo2026tabtracer}; \\ TabClaw~\cite{cheng2026tabclawinteractiveselfevolvingagent}; TaPERA~\cite{Zhao2024taperaenhancingfaithfulness_28}; \\ ST-Raptor~\cite{Tang2025raptorpoweredsemi_136};\\  FAMA~\cite{saeidi2026famafailureawaremetaagenticframework}; DataNarrative~\cite{Islam2024datanarrativeautomateddata_154}; Multi-Agent Visualization~\cite{wolter2025multiagentdatavisualizationnarrative}}, works-verification]
                    ]
                    [{Unstructured Data}, type-verification
                        [{UQE~\cite{Dai2024queryengineunstructured_164}; LongDS-Bench~\cite{xu2026longdsbenchfailurelonghorizonagentic}; DSAEval~\cite{sun2026dsaevalevaluatingdatascience}; MuDABench~\cite{li2026navigatinglargescaledocumentcollections}; UniDataBench~\cite{weng2026unidatabench};\\ Sanity Checks~\cite{rewolinski2026sanitychecksagenticdata}; Failure is Feedback~\cite{yun2026failurefeedbackhistoryawarebacktracking}}, works-verification]
                    ]
                ]
                [{Repair}, stage-repair, text width=6.6em
                    [{Structured Data}, type-repair
                        [{SagaLLM~\cite{chang2025sagallmcontextmanagement}; DART~\cite{yang2026dartsemanticrecoverability}; AgentFixer~\cite{mulian2026agentfixerfailuredetectionfix}; MERIT~\cite{wang2026learningretrieveduallevellongterm}; CausalFlow~\cite{bonagiri2026causalflowcausalattributioncounterfactual}}, works-repair]
                    ]
                    [{Semi-Structured Data}, type-repair
                        [{Pista~\cite{sabouri2026auditing}; Self-Healing Framework~\cite{jeong2026selfhealingframeworkreliablellmbased}; \\ DataCOPE~\cite{qiu2026unsupervisedskilldiscoveryagentic}; Fission-GRPO~\cite{zhang2026robusttoolusefissiongrpo};\\ Structured Reflection~\cite{su2026failure}; CoDA~\cite{chen2025codaagenticsystemscollaborative}; MatPlotAgent~\cite{Yang2024matplotagentmethodevaluation_152}}, works-repair]
                    ]
                    [{Unstructured Data}, type-repair
                        [{Doctor-RAG~\cite{jiao2026doctorragfailureawarerepairframework}; MOAR~\cite{wei2026multiobjectiveagenticrewritesunstructured}; \\ FinAcumen~\cite{guo2026finacumen}; Deep Search~\cite{sun2026deep}}, works-repair]
                    ]
                ]
            ]
        \end{forest}%
    }
    \caption{Data Agent taxonomy by workflow stage, data type, and execution signal.}
    \label{fig:data-agent-taxonomy}
\end{figure*}
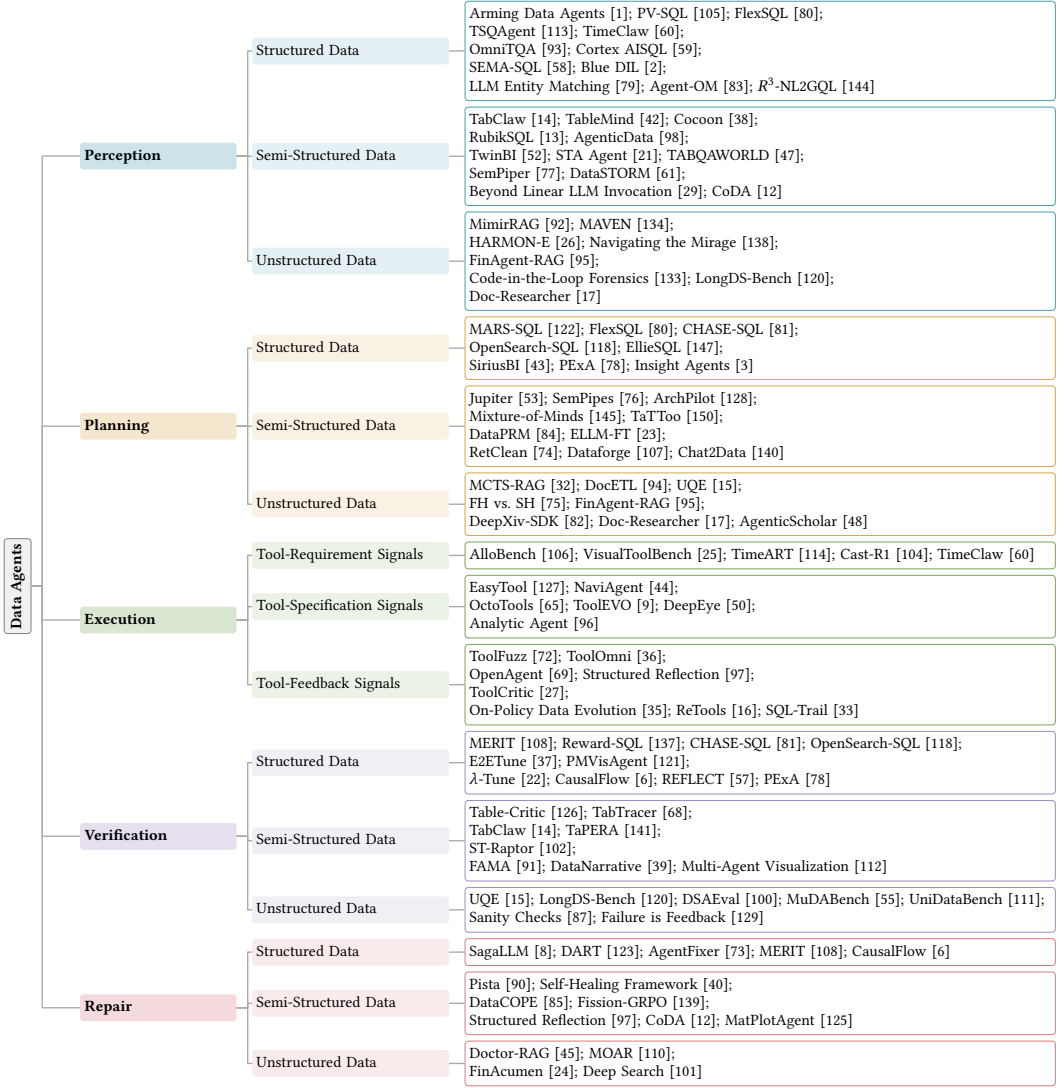

\subsection{Data Agents as AI Data Scientists}

To establish a rigorous foundation, we next define the concept of a Data Agent.

\begin{definition}[Data Agent]
A Data Agent is an LLM-driven system designed to execute data science tasks
through multi-step interaction with a persistent data environment. It actively
interacts with computational tools, e.g., Python sandboxes, visualization
libraries, and modeling frameworks. By observing results and iteratively refining its actions, the agent transforms data into verifiable analytical artifacts such as cleaned datasets, visualizations, reports, or trained models.
\end{definition}

\subsubsection{Operational Formalization of a Data Agent}

To formalize how a Data Agent operates, we model its execution as an evolving sequence of data states. Let:
\begin{itemize}
\item \( T \) be a data science task expressed in natural language;
\item \( \mathcal{D} \) be the data environment as defined in Section~2.2, with \( \mathcal{D}_0 \) denoting its initial state;
\item \( \mathcal{E} \) be the execution environment exposing an action space \( \mathcal{A}_{\mathcal{E}} \), e.g., calling a tool to search information online and an observation space \( \mathcal{O}_{\mathcal{E}} \), e.g., compiler errors or statistical summaries;
\item \( M \) be the underlying LLM that acts as the reasoning core.
\end{itemize}

A Data Agent maps this initial setting to a finite trajectory of actions, observations, and evolving data states, culminating in a final artifact \( R \):
$$
A(T, \mathcal{D}_0, \mathcal{E}, M) \rightarrow \big((a_1, o_1, \mathcal{D}_1), (a_2, o_2, \mathcal{D}_2), \ldots, (a_n, o_n, \mathcal{D}_n), R\big)
$$
where \( a_i \in \mathcal{A}_{\mathcal{E}} \) is a concrete data-centric action at step \( i \), \( \mathcal{D}_i \) represents the updated, underlying data state after applying \( a_i \), \( o_i \in \mathcal{O}_{\mathcal{E}} \) is the observation returned by \( \mathcal{E} \) which serves as a partial view of \( \mathcal{D}_i \), and \( h_i = ((a_1, o_1, \mathcal{D}_1), \ldots, (a_{i-1}, o_{i-1}, \mathcal{D}_{i-1})) \) is the execution history that conditions the agent's next decision.

The scope of Data Agents includes data science tasks that span two broad families: (1) \textit{data analytics} tasks, including data profiling, visualization, and data manipulation, e.g., spreadsheet editing; and (2) \textit{ML-oriented data} tasks, including feature engineering, model selection, and model training. In this survey, we use the term Data Agents to refer to systems that interact with a data environment through multiple steps, where subsequent actions can depend on intermediate observations and execution outcomes. This distinguishes them from systems restricted to one-shot prediction, or generation without iterative interaction with the underlying data environment.

\subsubsection{Handling Silent Failures with a Workflow Harness}
A dangerous class of failure in data analysis is the \textit{silent failure}, where a Data Agent terminates a task without raising explicit errors, yet produces semantically incorrect or incomplete outputs. Other manifestations include silent timeouts without meaningful outputs, or downstream errors that surface long after the initial execution.

Because the real-world data environment \( \mathcal{D} \) is inherently imperfect, often plagued by missing metadata \( \Sigma_j \), implicit properties \( \Pi_j \), and heterogeneous data types \( \tau_j \), mistakes such as grounding on the wrong evidence or exceeding resource limits are common. Therefore, the challenge is not only to enhance the agent's capabilities but also to govern its decisions and resource consumption through a comprehensive workflow harness. We distinguish the workflow from the workflow harness. The workflow is the
task-specific trajectory generated by the Data Agent as it interacts with the data environment, i.e., the evolving sequence of actions, observations, and data states formalized above. The workflow harness, in contrast, is the control structure that governs how this trajectory is constructed, monitored, and revised. Rather than prescribing a fixed sequence of
operations, it determines how the agent interprets the current data state,
selects subsequent actions, evaluates their outcomes, and responds when the
workflow deviates from the task objective. As detailed in Section~3, we decompose this control structure into five functional stages: perception, planning, execution, verification, and repair.

Traditional workflow orchestrators, such as Airflow, execute developer-specified task graphs. In contrast, a Data Agent selects, evaluates, and revises analytical actions as the data state evolves, allowing it to detect and respond to silent failures during execution.

\section{The Workflow Harness of AI Data Scientists}

Building on the preceding formulation, this section examines the five
functional stages through which the workflow harness governs a Data Agent's
task trajectory: perception, planning, execution, verification, and repair.
Rather than forming a fixed sequence that is executed once from beginning to
end, these stages are invoked as needed according to the evolving data
state, allowing the agent to revisit earlier stages and iterate between them
throughout task execution.

Within each stage, we organize existing methods into technical routes
according to the recurring mechanisms through which they fulfill the
stage-specific functional objective. These routes are complementary rather
than mutually exclusive, and a Data Agent may instantiate multiple routes.

The overall workflow harness is illustrated in Figure~\ref{fig:Workflow},
and we examine five research questions: how does the agent build a
structured understanding of the current data environment state?
(Section~\ref{sec:perception}); how does it choose a feasible and near-optimal
plan under uncertainty? (Section~\ref{sec:planning}); how does it execute the
formulated plan through computational tools? (Section~\ref{sec:execution});
how does it routinely validate its ongoing progress to locate silent failures?
(Section~\ref{sec:verification}); and how does it repair these failures and
retain the learned corrections as reusable lessons for future tasks?
(Section~\ref{sec:repair}).
\begin{figure*}[htbp]
    \centering
    \includegraphics[width=\linewidth]{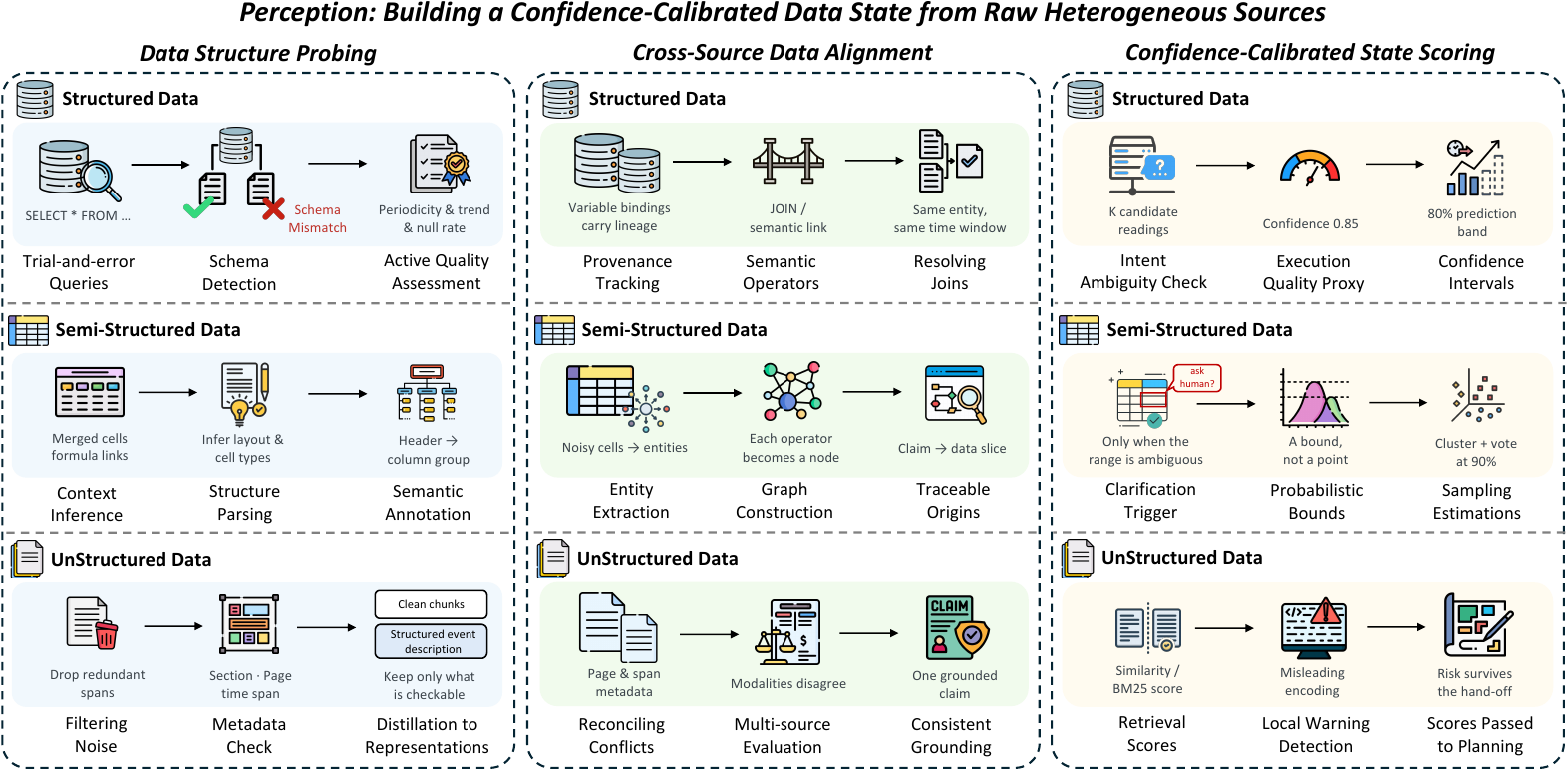}
    \caption{Technical routes for perception across heterogeneous data
environments: Data Structure Probing, Cross-Source Data Alignment,
and Confidence-Calibrated State Scoring.}
    \label{fig:cognitive-capability-framework}
\end{figure*}

\subsection{Perception}
\label{sec:perception}
Perception is the entry point of the Data Agent workflow. Early systems simply injected a schema or table into the prompt. Modern systems have moved beyond this: they probe data sources, connect data across modalities, and attach confidence to what they perceive. We group these advances into three technical routes, summarized in Figure~\ref{fig:cognitive-capability-framework}. \textbf{Data structure probing} inspects the organization of data and samples values to ground task intent in data objects. \textbf{Cross-source data alignment} connects extracted data across heterogeneous data sources, e.g., databases, documents, and dashboards. \textbf{Confidence-calibrated state scoring} attaches confidence scores to perceived data states, enabling downstream stages to identify which assumptions are fragile.

\subsubsection{Data Structure Probing}

Data structure probing builds and progressively refines the agent's understanding of the data environment by issuing exploratory queries or inspecting structures and sample values~\cite{agarwal2026arming}. Rather than treating the perceived state as a static snapshot, probing iteratively updates it as new structural or statistical evidence becomes available.

\textbf{Structured Data.} In relational databases, probing typically relies on trial-and-error execution to uncover schema properties and validate tentative assumptions. For instance, PV-SQL~\cite{tian2026pvsqlsynergizingdatabaseprobing} constructs tentative SQL queries and observes execution errors to detect schema mismatches. Taking this step further, FlexSQL~\cite{pham2026flexsql} generates multiple SQL variants and uses their execution outcomes to refine subsequent queries. Beyond relational tables, probing in time-series data shifts from schema exploration toward statistical inspection. Approaches differ in how they determine what to inspect: TSQAgent~\cite{wu2026tsqagentratingtimeseries} employs a unified perceiver, optimized via GRPO, to actively select relevant quality dimensions, whereas TimeClaw~\cite{liu2026timeclawtimeseriesaiagent} formulates probing as a routing problem, using supervised learning to determine which exploratory queries should be applied. Overall, structured data probing is evolving from passive schema inspection toward dynamic, feedback-driven exploration of both structural and statistical properties.

\textbf{Semi-Structured Data.} In spreadsheets, notebooks, and other semi-structured environments, probing focuses on uncovering implicit structure and inferring the local context needed to interpret the data correctly. At the structural level, TabClaw~\cite{cheng2026tabclawinteractiveselfevolvingagent} explicitly parses properties such as merged cells and formula dependencies to identify safe modification zones, whereas TableMind~\cite{jiang2026tablemind} analyzes complex table structures through a perceiver trained with a two-stage fine-tuning paradigm. Beyond structural inspection, recent work also probes statistical and contextual cues to infer latent semantics. Cocoon~\cite{Huang2024cocoonsemantictable_122}, for example, provides column statistics to an LLM to determine whether observed patterns reflect domain-specific semantics. Extending this idea to broader and more heterogeneous data environments, RubikSQL~\cite{chen2025rubiksqllifelonglearningagentic} and AgenticData~\cite{sun2025agenticdata} profile data sources through agentic rule mining. Overall, semi-structured data probing progressively turns implicit layouts, local dependencies, and statistical cues into an explicit understanding of the data context.

\textbf{Unstructured Data.} In unstructured environments such as documents and videos, probing focuses on selecting relevant evidence and transforming raw inputs into representations that preserve the context needed for downstream reasoning. For textual data, agents may inspect metadata and section boundaries to determine which chunks should enter the perceived data state. MimirRAG~\cite{samuelsen2026mimirragmultiagentragframework}, for example, verifies metadata consistency before passing selected chunks to subsequent stages. High-dimensional modalities such as video require more substantial transformation. MAVEN~\cite{zhang2026mavenmultistageagenticannotation} probes video evidence by constructing multiscale spatiotemporal event descriptions, transforming raw frames into structured representations. Consequently, unstructured data probing acts as a selective distillation process that reduces large and noisy inputs while preserving the information needed to construct a useful perceived state.
\subsubsection{Cross-Source Data Alignment}

Cross-source data alignment connects semantically related information across heterogeneous data sources, allowing the agent to construct a coherent perceived state from otherwise fragmented observations. The core challenge is that corresponding information may be expressed under different schemas, formats, granularities, or representations. Alignment therefore requires not only identifying semantic correspondences across sources, but also preserving the provenance that links aligned information back to its original data.

\textbf{Structured Data.} In structured data environments, alignment primarily concerns preserving semantic correspondences across complex schemas and routing information to the appropriate sources. OmniTQA~\cite{shahbazi2026omnitqacostawarehybridquery} employs a semantic operator to track mappings between query variables and external sources, propagating these mappings to dependent operations involving the grounded variables. Cortex AISQL~\cite{liskowski2026cortex} extends this paradigm by embedding such semantic operators directly into SQL queries. When schemas become more complex, alignment further requires semantic matching and source routing. SEMA-SQL~\cite{lin2026semasqltraditionalrelationalquerying} introduces neural semantic matching components to align variable representations and resolve naming mismatches across schemas. Complementing this, the Blue Data Intelligence Layer~\cite{aminnaseri2026bluedataintelligencelayer} provides a declarative DataPlanner that routes variables in decomposed subtasks to their appropriate data sources. Overall, structured data alignment maintains explicit links between derived results and the source fields from which they originate.

\textbf{Semi-Structured Data.} In semi-structured data environments, alignment connects locally extracted elements to higher-level structural representations while preserving their original context. Within tabular layouts, this process often begins by resolving ambiguities in individual cells. Semantic table annotation systems~\cite{geng2025llmagentbasedcomplexsemantic} extract entities from noisy table cells and align them using named entity recognition and rule-based matching. Building on this, TABQAWORLD~\cite{kwok2026tabqaworldoptimizingmultimodalreasoning} binds each entity to its corresponding column group and row hierarchy, situating locally extracted entities within the surrounding table structure. Beyond individual tables, alignment can extend to graph structures that preserve traceability across intermediate representations. SemPiper~\cite{ovcharenko2026sempiperinteractivecodesynthesis} records data transformations through a dataflow graph, allowing downstream features to remain traceable to their semi-structured origins. Scaling this graph-based organization to more complex domains, DataSTORM~\cite{liu2026datastormdeepresearchlargescale} connects extracted concepts into a hierarchical graph in which each node can be traced back to specific data slices. Consequently, semi-structured data alignment progressively connects locally extracted entities to broader structural representations.

\textbf{Unstructured Data.} In unstructured data environments, the central problem therefore shifts from matching predefined fields to reconciling semantically related, and potentially conflicting, evidence across heterogeneous inputs. HARMON-E~\cite{gupta2025harmonehierarchicalagenticreasoning} aligns evidence across heterogeneous clinical records by merging information about the same entity from multiple documents and resolving duplicated or conflicting attributes using domain-specific rules. The Deception-Aware GRPO mechanism~\cite{zhang2026navigatingmiragedualpathagentic} further evaluates evidence from diverse sources and halts extraction when conflicts cannot be reliably resolved. Overall, unstructured data alignment seeks to establish coherent correspondences across heterogeneous evidence while preventing unresolved conflicts from being propagated into the perceived data state.

\subsubsection{Confidence-Calibrated State Scoring}

Data structure probing and cross-source data alignment construct the perceived data state, but neither directly indicates how much that state should be trusted. This distinction is important because a perceived state may appear structurally valid while still relying on uncertain matches, incomplete evidence, or fragile assumptions. Confidence-calibrated state scoring therefore attaches reliability signals to the perceived data state, allowing downstream workflow stages to distinguish well-supported information from uncertain information and determine whether the current state is sufficient to proceed or requires further probing.

\textbf{Structured Data.} In structured data environments, confidence can often be estimated from explicit statistical or model-based evidence because individual data objects and their relationships are relatively well defined. Entity-matching methods~\cite{Peeters2025entitymatchingusing_133}, for example, estimate the reliability of candidate matches from model probabilities or disagreement across predictions. Such signals indicate whether a particular correspondence should be incorporated into the perceived data state or treated as uncertain. In temporal data, confidence can instead reflect the reliability of inferred data properties. TSQAgent~\cite{wu2026tsqagentratingtimeseries} evaluates time-series quality along selected dimensions and associates its judgments with explicit confidence scores, allowing uncertain quality assessments to remain distinguishable from more reliable ones. Thus, structured data scoring converts statistical evidence associated with individual data objects or inferred properties into explicit reliability signals.

\textbf{Semi-Structured Data.} In semi-structured environments, confidence scoring is complicated by the fact that important structural information is often implicit rather than directly encoded. Ambiguities in headers, cell ranges, or table organization can therefore affect how the agent interprets the surrounding data context. TabClaw~\cite{cheng2026tabclawinteractiveselfevolvingagent} handles such uncertainty by treating unresolved structural ambiguity as a signal that the current perceived state is insufficient, triggering clarification before subsequent operations are performed. Beyond Linear LLM Invocation~\cite{hou2026linearllminvocationefficient} addresses the same problem at a larger scale by estimating the reliability of table-level decisions from sampled records. It groups similar records, evaluates representative samples, and further examines groups whose decisions remain uncertain rather than applying the LLM uniformly to every record. Consequently, semi-structured data scoring determines whether inferred structure is reliable enough for subsequent workflow stages while avoiding unnecessary inspection of stable data.

\textbf{Unstructured Data.} In unstructured environments, reliability signals are often available during perception but can become detached from the perceived data state as evidence is transformed and passed downstream. Document retrieval systems~\cite{samuelsen2026mimirragmultiagentragframework,shu2026agenticretrievalaugmentedgenerationfinancial}, for example, assign relevance scores when selecting evidence, providing a local indication of how strongly each retrieved item supports the current state. Visual analysis systems~\cite{zhang2026navigatingmiragedualpathagentic,zhang2026codeintheloopforensicsagentictool} similarly identify suspicious visual evidence before it is incorporated into subsequent reasoning. The difficulty is that these local reliability signals may no longer remain explicit once the selected evidence has been summarized or transformed into a new state representation. LongDS-Bench~\cite{xu2026longdsbenchfailurelonghorizonagentic} further illustrates how errors introduced into an early state can propagate through long-horizon analytical workflows. Therefore, confidence-calibrated state scoring in unstructured environments must not only estimate the reliability of local evidence, but also preserve that reliability as part of the perceived data state when it moves to downstream workflow stages.

\paragraph{Technical Route Summary.}
The three perception routes address distinct but complementary aspects of constructing a reliable perceived data state. Data structure probing asks, ``what does the data environment look like?'' It actively inspects the data to uncover the structures, properties, and context needed to form the perceived state. Cross-source data alignment asks, ``which data objects belong together?'' It establishes semantic correspondences across sources while preserving their links to the original data. Confidence-calibrated state scoring asks, ``how much should this perceived state be trusted?'' It associates the state with reliability signals so that later workflow stages can distinguish well-supported information from uncertain assumptions.

Technically, these routes rely on distinct mechanisms to construct and refine the perceived data state. Probing is exploration-driven, using trial-and-error queries, structural parsing, and statistical profiling to reveal previously unknown properties of the data environment. Alignment is semantics- and provenance-driven, relying on semantic matching to establish correspondences and graph-based representations to preserve traceability across sources. Scoring is uncertainty-driven, deriving reliability signals from model confidence, sampling-based estimation, or disagreement, while preserving these signals as the perceived state moves to downstream workflow stages.

\begin{figure*}[htbp]
    \centering
    \includegraphics[width=\linewidth]{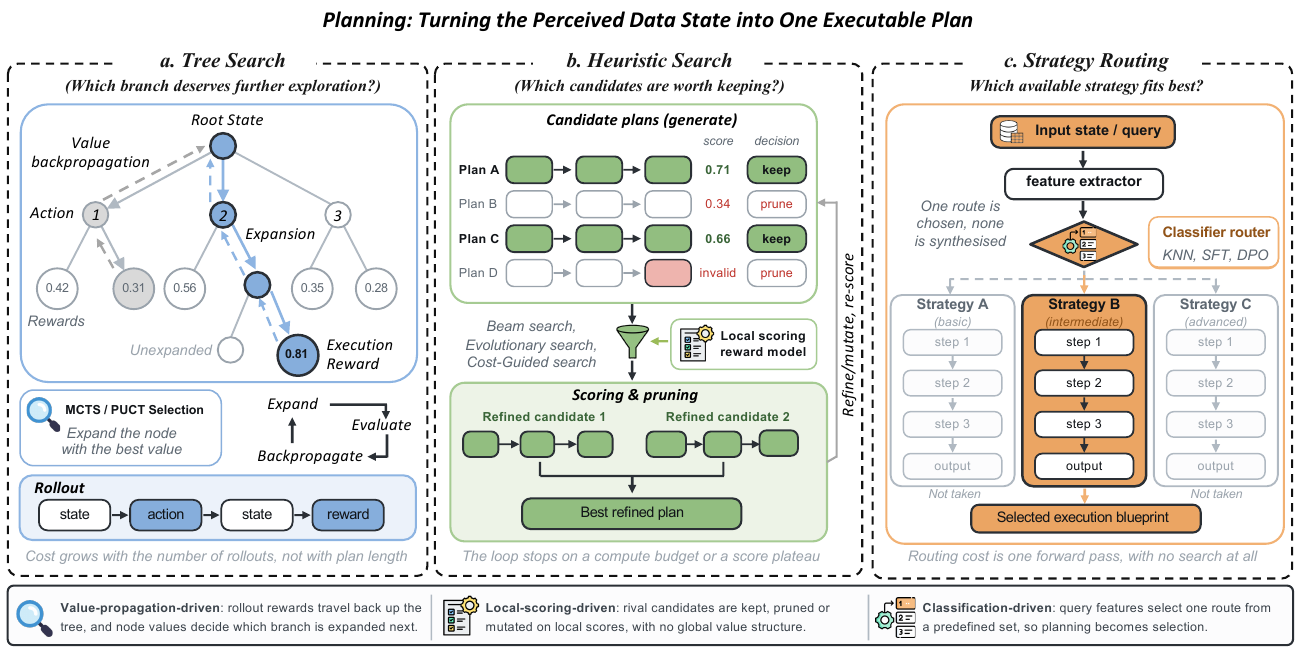}
    \caption{Technical routes for planning: Tree Search, Heuristic Search,
and Strategy Routing.}
    \label{fig:planning-routes}
\end{figure*}


\subsection{Planning}
\label{sec:planning}

Planning translates the perceived data state into an executable blueprint. For Data Agents, a reliable plan must connect high-level task intent with concrete operations over the current data environment. Existing approaches can be broadly categorized into three technical routes: \textbf{tree search}, \textbf{heuristic search}, and \textbf{strategy routing}. These planning routes are summarized in Figure~\ref{fig:planning-routes}. Tree search and heuristic search construct plans by actively exploring candidate paths, whereas strategy routing selects among predefined strategies according to the current perceived state.

\subsubsection{Tree Search}

Tree search formulates planning as explicit exploration over a branching decision space. Rather than committing to a single autoregressive plan, the agent maintains multiple candidate trajectories and allocates further exploration according to their estimated values. Methods such as Monte Carlo Tree Search (MCTS) and Predictor Upper Confidence Trees (PUCT) further propagate downstream outcomes, allowing later execution to refine preceding choices.

\textbf{Structured Data.} In structured data environments, tree search appears more often as a mechanism for generating planning trajectories than as a persistent inference-time controller. The explicit structure of relational and temporal data already provides strong constraints on feasible operations, reducing the need to maintain a large search tree throughout execution. MARS-SQL~\cite{yang2026marssqlmultiagentreinforcementlearning} follows this pattern by using MCTS-style rollouts to generate planning, coding, and answering trajectories for GRPO optimization. Search is therefore used to expose diverse decision paths during training, while the learned policy performs the subsequent inference-time planning. This setting illustrates one role of tree search in structured environments: converting a well-defined action space into diverse trajectories from which better planning behavior can be learned.

\textbf{Semi-Structured Data.} In semi-structured environments, tree search is used more directly to explore alternative planning paths because the underlying structure is only partially specified and intermediate decisions can often still be evaluated. Jupiter~\cite{li2026jupiter} uses MCTS to collect trajectories and derives a value model from normalized MCTS Q-values. During inference, this learned value signal guides further tree search, allowing exploration to concentrate on promising branches. SemPipes~\cite{ovcharenko2026sempipesoptimizablesemantic} applies the same search principle to semantic operator synthesis, where candidate programs form the search tree and downstream validation performance determines which branches should be expanded. ArchPilot~\cite{yuan2025archpilotproxyguidedmultiagentapproach} extends value-guided search to ML engineering by using inexpensive proxy scores to guide architecture exploration and restarting the tree when previous estimates become unreliable. Mixture-of-Minds~\cite{zhou2025mixture} also uses MCTS in table reasoning, generating candidate coding and answering trajectories that serve as pseudo-gold data for subsequent reinforcement learning. Across these settings, tree search supports either the selection of promising analytical paths or the collection of trajectories for improving future planning. Across these settings, tree search becomes practical when intermediate states can be compactly represented.

\textbf{Unstructured Data.} In unstructured data environments, direct tree search is less straightforward because raw documents or multimodal evidence do not naturally expose discrete states with readily computable values. Recent approaches address this problem by discretizing the planning process rather than searching directly over the raw data space. MCTS-RAG~\cite{hu2025mctsrag}, for example, defines retrieval-oriented actions such as query decomposition and active retrieval as explicit branches of the search tree, then uses LLM-based confidence estimates to score intermediate states. By converting open-ended reasoning over unstructured evidence into discrete, scoreable decisions, such approaches enable value-guided tree search without explicit structural constraints.

\subsubsection{Heuristic Search}

Heuristic search explores alternative plans without explicitly maintaining a global search tree. Instead of recursively expanding and backpropagating values through tree nodes, it iteratively generates candidate trajectories and retains promising ones according to task-specific scoring signals. Common instantiations include beam search, evolutionary search, and cost-guided local search. By avoiding explicit tree construction, heuristic search adapts to large or irregular action spaces, but its exploration depends more on the quality of the local heuristic.

\textbf{Structured Data.} In structured data environments such as SQL generation, heuristic search commonly follows a generate-and-refine pattern in which multiple candidate solutions are produced and progressively filtered using execution or model-based feedback. FlexSQL~\cite{pham2026flexsql} encourages diversity among sampled SQL candidates and uses majority voting to select among their execution outcomes. CHASE-SQL~\cite{Pourreza2025chasemultipath_83} broadens the candidate space through multiple reasoning paths and applies preference-based scoring to retain stronger solutions. OpenSearch-SQL~\cite{Xie2025opensearchenhancingtext_215} instead couples candidate refinement with dynamic few-shot retrieval, repeatedly revising SQL generation as more relevant examples are retrieved. These approaches therefore replace single-pass decoding with iterative candidate exploration, where execution results, preference scores, or retrieval evidence determine which solutions remain under consideration.

\textbf{Semi-Structured Data.} In semi-structured environments, heuristic search is often used to control the growth of partial analytical trajectories through beam-based pruning or population-based optimization. TaTToo~\cite{zou2026tattoo} employs beam search and uses reward scores to discard weak partial trajectories while preserving promising ones for further expansion. DataPRM~\cite{qiu2026rewarding} similarly introduces process-level rewards as intermediate pruning signals, allowing search to concentrate on trajectories that remain plausible before a final outcome is available. ELLM-FT~\cite{Gong2025evolutionarylargelanguage_44} follows a different search mechanism by maintaining a population of feature-transformation sequences and using downstream validation performance as the fitness signal for selection. Despite their different implementations, these methods share the same principle: intermediate quality signals are used to repeatedly narrow a large candidate space without constructing an explicit global search tree.

\textbf{Unstructured Data.} In unstructured data environments, evaluating intermediate candidates is often expensive because each alternative may require additional retrieval, semantic processing, or downstream execution. Heuristic search therefore tends to favor localized refinement and cost-aware evaluation rather than broad candidate expansion. DocETL~\cite{Shankar2025docetlagenticquery_198}, for example, applies rewrite-based local search to document-processing pipelines and accepts operator modifications only when sampled downstream evaluation indicates an improvement. UQE~\cite{Dai2024queryengineunstructured_164} similarly uses sample-guided cost estimates to compare retrieval operators before committing to full execution. Such methods restrict expensive evaluation to a small set of promising candidates, making search tractable when exhaustive exploration over unstructured evidence would be impractical. Recent results further suggest that expanding the planning space indiscriminately can be counterproductive when additional search does not align with the structure of the task~\cite{otani2026agents}.

\subsubsection{Strategy Routing}

Strategy routing selects among a predefined set of planning strategies according to the current perceived data state and task requirements. Unlike search-based planning, which constructs and evaluates new candidate paths during planning, routing operates over an existing strategy space and determines which strategy should be activated. The selected strategy may change as new information arrives, but the strategy itself is predefined.

\textbf{Structured Data.} In structured data environments, routing commonly matches task complexity to an appropriate analytical strategy or model. EllieSQL~\cite{zhu2025elliesql}, for example, trains complexity-aware routers using KNN, SFT, or DPO variants to direct Text-to-SQL queries toward basic, intermediate, or advanced generation models. SiriusBI~\cite{Jiang2025siriusbicomprehensivepowered_159} similarly selects between one-step and two-step SQL generation according to the complexity of the query. In these settings, routing avoids exploring multiple candidate plans by estimating the requirements of the current task and directly activating a suitable predefined strategy.

\textbf{Semi-Structured Data.} In semi-structured environments, routing often determines which retrieval or transformation strategy best fits the current data context and operational constraints. RetClean~\cite{Naeem2024retcleanretrievalbased_48} selects among no retrieval, public retrieval, and private retrieval according to the information required and its privacy constraints. Dataforge~\cite{wang2026dataforgeagenticplatformautonomous} organizes this decision hierarchically, first routing a task to a broad feature-processing family before selecting more specific operations within that family. Routing can also be revised when the current strategy proves insufficient: Chat2Data~\cite{Zhao2024chat2datainteractivedata_225} shifts from single-turn to multi-round retrieval when user feedback indicates that additional information is needed. Thus, semi-structured routing maps the evolving data context to a predefined strategy space without searching over alternative trajectories.

\textbf{Unstructured Data.} In unstructured data environments, routing frequently controls the depth and cost of information acquisition. Financial RAG agents~\cite{shu2026agenticretrievalaugmentedgenerationfinancial} use model confidence to choose between single-pass and iterative retrieval, increasing retrieval depth only when the current evidence is insufficient. DeepXiv-SDK~\cite{qian2026deepxivsdkagenticdatainterface} routes scientific literature queries to different access layers according to resource and cost constraints. In these settings, routing extends planning by incorporating confidence and resource constraints when selecting predefined retrieval strategies.

\paragraph{Technical Route Summary.}
The three planning routes address distinct aspects of selecting an executable path from the current perceived data state. Tree search asks, ``Which branches deserve further exploration?'' by constructing an explicit decision tree and using estimated node values to guide expansion. Heuristic search asks, ``Which candidates are worth keeping?'' by iteratively generating alternatives and retaining promising ones according to local quality signals. Strategy routing asks, ``Which available strategy fits the current state?'' by selecting from a predefined strategy space rather than constructing new candidate paths.

Technically, the three routes differ in how planning alternatives are represented and selected. Tree search is \textit{value-propagation-driven}: methods such as MCTS and PUCT maintain an explicit branching structure, estimate the value of intermediate states, and propagate downstream outcomes to earlier decisions. Heuristic search is \textit{local-scoring-driven}: beam search, evolutionary search, and cost-guided evaluation repeatedly narrow the candidate space without keeping a global tree. Strategy routing is \textit{state-to-strategy mapping-driven}: learned routers, confidence scores, or task constraints map the current state to a predefined strategy, with rerouting when the state changes.


\begin{figure*}[htbp]
    \centering
    \includegraphics[width=\linewidth]{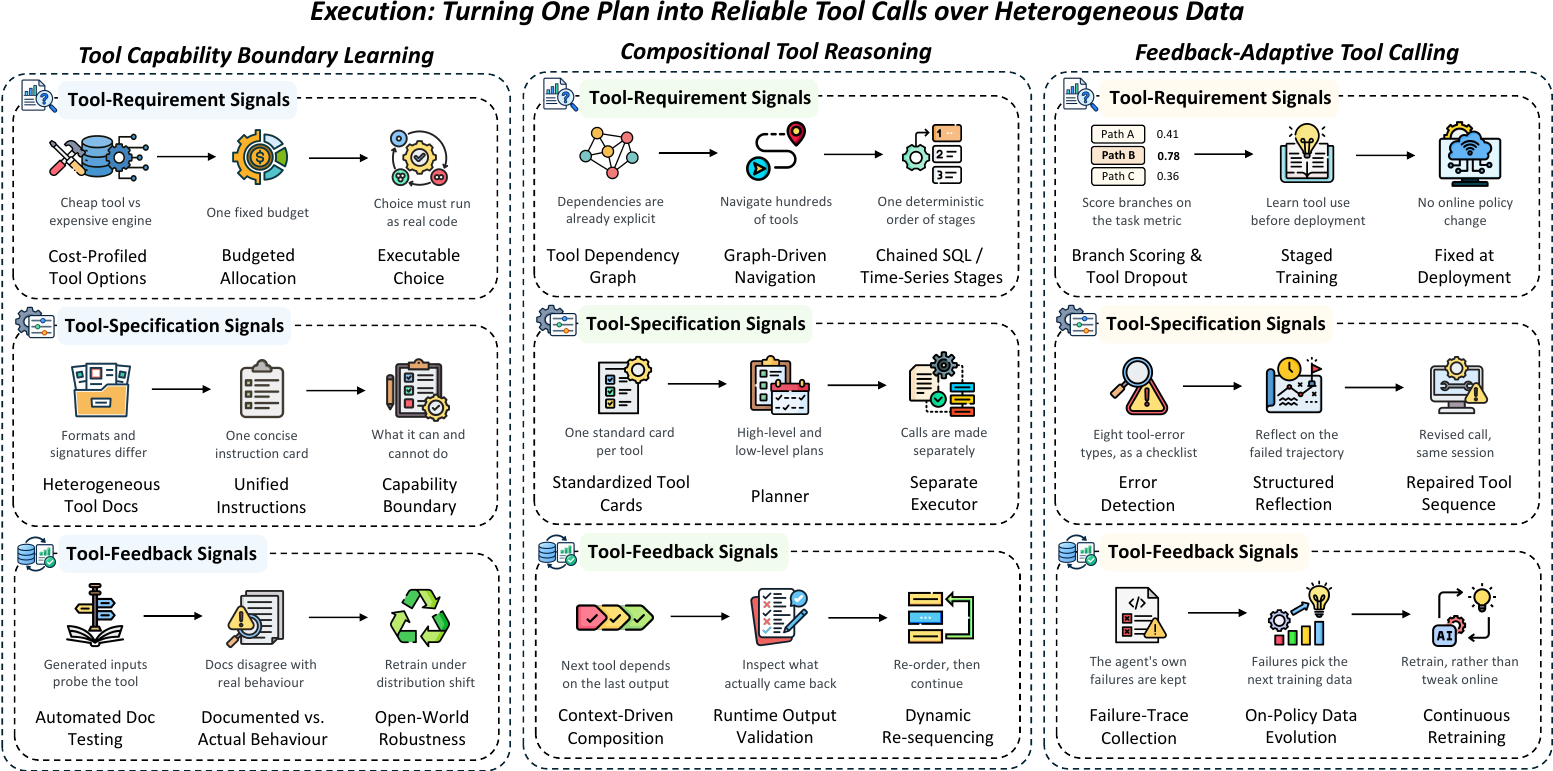}
    \caption{Technical routes for execution: Tool Capability Boundary Learning,
Compositional Tool Reasoning, and Feedback-Adaptive Tool Calling.
}
    \label{fig:execution-routes}
\end{figure*}

\subsection{Execution}
\label{sec:execution}

Execution turns a planned analytical step into a concrete interaction with the
data environment. In a Data Agent, tools serve as the executable interface
between the analytical plan and the current data state. Reliable execution
therefore depends on three sources of signals:
\textbf{Tool-Requirement Signals}, which specify what the current analytical
state requires from the next operation;
\textbf{Tool-Specification Signals}, which describe the capabilities,
interfaces, and constraints of available tools; and
\textbf{Tool-Feedback Signals}, which capture evidence returned by actual tool
execution. These signals are complementary and may jointly inform a single
tool-use decision.

Based on how these signals support tool-mediated interaction with the data
environment, existing approaches are organized into three technical routes:
\textbf{Tool Capability Boundary Learning},
\textbf{Compositional Tool Reasoning}, and
\textbf{Feedback-Adaptive Tool Calling}, as shown in
Figure~\ref{fig:execution-routes}.

\subsubsection{Tool Capability Boundary Learning}

Tool capability boundary learning aims to establish an accurate understanding
of what a tool can and cannot reliably do. A capability boundary specifies the
range of operations and inputs that a tool is designed to support. For Data
Agents, the relevant question is not merely whether a tool is available, but
whether its capability matches the operation required by the current analytical
state.

\textbf{Tool-Requirement Signals.}
The current analytical objective determines the capability that must be
provided by a tool. These requirements can include the type of operation to be
performed, the expected form of its result, and operational constraints such as
cost or resource budgets. AlloBench~\cite{wang2026allobenchmeasuringonline}
isolates this requirement--capability matching problem through paired allocation
tasks. Frontier models can make near-optimal allocation decisions in an abstract
textual setting, yet often fail to preserve the same decisions when they must
instantiate them through executable code. Although studied in a generic tool
allocation setting, this mismatch is directly relevant to Data Agents: correctly
identifying the required capability does not guarantee that the agent can
realize it as a valid operation over the current data environment.

\textbf{Tool-Specification Signals.}
Requirement matching depends on an explicit representation of what candidate
tools support. Such information is commonly exposed through documentation,
function signatures, capability descriptions, or other tool metadata, but these
representations can be lengthy, heterogeneous, or incomplete. EasyTool~
\cite{yuan2025easytool} addresses this problem by rewriting diverse tool
documentation into concise and standardized instructions, reducing descriptive
variation while preserving the operational information needed to determine when
a tool should be used. In Data Agents, such specification signals provide the
bridge between an analytical requirement expressed over the current data state
and the operations exposed by the tool environment.

\textbf{Tool-Feedback Signals.}
Declared capabilities do not always coincide with observed behavior.
Consequently, execution itself can provide evidence for refining the estimated
capability boundary. ToolFuzz~\cite{milev2025toolfuzzautomatedagent} generates
calling tasks to probe tools and expose discrepancies between documented and
actual behavior. ToolOmni~\cite{huang2026toolomnienablingopenworld} further
couples tool retrieval with downstream task performance, allowing empirical
execution quality to influence which tools are considered useful within a large
candidate set. OpenAgent~\cite{lv2026agentsgeneralizeopenworld} shows that
previously learned tool-use knowledge can also become unreliable under domain
distribution shifts and improves robustness through perturbation-augmented
fine-tuning. Together, these approaches treat execution evidence as a means of
correcting an agent's estimate of tool capability rather than assuming that
tool specifications are universally reliable.

Overall, capability boundary learning can be viewed as an alignment problem:
tool-requirement signals specify what the current analytical state needs,
tool-specification signals describe what candidate tools are expected to
support, and tool-feedback signals refine this alignment when observed behavior
deviates from the specification.

\subsubsection{Compositional Tool Reasoning}

Many analytical tasks cannot be completed by a single tool call. Instead,
multiple operations must be connected so that each intermediate result provides
the information required by subsequent steps. Compositional tool reasoning
therefore determines how tool-mediated operations should be assembled into a
coherent execution sequence while preserving the evolving analytical state.

\textbf{Tool-Requirement Signals.}
Composition begins from the analytical requirements of the current state.
Rather than selecting tools independently, the agent must determine which
sub-operation is needed next and what intermediate result must be produced for
the remaining workflow. This dependency is especially visible in multimodal
data analysis. VisualToolBench~\cite{guo2025seeingevaluatingmultimodalllms}, e.g., requires agents to manipulate visual inputs through multiple
tool-mediated transformations and integrate the resulting evidence with
general-purpose tools. A crop or extraction operation changes the
available evidence and therefore changes the requirement imposed on the next
tool call. Composition is thus grounded in the evolving analytical state rather
than in a fixed tool sequence.

\textbf{Tool-Specification Signals.}
Once the required sub-operations are identified, the agent must understand how
available tools can be connected. NaviAgent~\cite{jiang2026naviagent} makes such
relations explicit through a navigation graph whose nodes and edges encode
dependencies among tools. Its Tool World Navigation Model allows the agent to
traverse these relations and construct executable invocation sequences over
large tool sets. OctoTools~\cite{lu2026octotools} instead represents
heterogeneous tools through standardized tool cards and combines them with
hierarchical planning. A high-level plan identifies the analytical objective
and relevant tools, while low-level planning determines the next tool, the
required information, and the corresponding sub-goal before the executor issues
the actual call. These approaches show how tool specifications can expose the
functional and dependency information needed to translate analytical
requirements into executable compositions.

\textbf{Tool-Feedback Signals.}
Intermediate tool outputs also participate in composition because the result of
one operation becomes part of the analytical state from which the next
requirement is derived. This role of feedback differs from explicit error
correction: here, an execution result primarily supplies content or state
information needed to continue the tool chain. In iterative visual analysis,
for example, the output of one transformation may reveal which subsequent
operation is useful, making the execution sequence dependent on what the
previous call actually returned~\cite{guo2025seeingevaluatingmultimodalllms}.
Thus, compositional reasoning links tools not only through predefined interface
dependencies, but also through the intermediate data states produced during
execution.

In this route, the three signals play complementary roles: analytical
requirements determine what intermediate operation is needed, tool
specifications determine which operations can be connected, and execution
feedback updates the state from which the next requirement is formed.

\subsubsection{Feedback-Adaptive Tool Calling}

Feedback-adaptive tool calling focuses on revising tool-use behavior when
execution evidence indicates that the current choice, parameterization, or
calling strategy is ineffective. Unlike compositional tool reasoning, where an
intermediate output is primarily used to continue a dependent sequence,
feedback adaptation interprets execution evidence evaluatively: the agent asks
whether the current tool use has actually advanced the analytical objective and,
if not, how it should be changed.

\textbf{Tool-Requirement Signals.}
Execution feedback is meaningful only relative to the analytical requirement
that the tool was intended to satisfy. In data analysis tasks, task-specific
metrics and properties of the current data state can therefore define what
counts as successful tool use. TimeART~\cite{wu2026timeartagentictimeseries}
trains time-series agents over staged tool-use trajectories so that execution
experience shapes the tool-use behavior learned before deployment.
TimeClaw~\cite{liu2026timeclawtimeseriesaiagent} further evaluates exploratory
tool trajectories with task-specific metrics and introduces task-aware tool
dropout to reduce reliance on tool choices that produce suboptimal analytical
results. These methods ground adaptation in the requirements of the underlying
data task rather than treating tool success as an abstract property of the call
itself.

\textbf{Tool-Specification Signals.}
When feedback indicates that a tool call is inadequate, the agent must determine
which alternative use of the available tool space remains feasible. Tool
specifications therefore constrain the space of possible corrections.
ToolEVO~\cite{chen2025learning} studies this problem in environments where the
available tools and their interfaces evolve over time. Agents actively explore
the changed environment, reflect on feedback, and update their behavior as
previous tool-use knowledge becomes outdated. Here, adaptation requires not
only recognizing failure but also revising the agent's understanding of the
currently available tool capabilities.

\textbf{Tool-Feedback Signals.}
The most direct adaptation signal comes from the execution trace itself.
Structured-reflection approaches deliberately expose agents to unsuccessful
tool trajectories so that they learn to diagnose failed calls and transform
them into successful executions~\cite{su2026failure}. ToolCritic~
\cite{hamad2025toolcriticdetectingcorrecting} formalizes this process by
detecting eight categories of tool-calling errors and providing targeted
feedback that guides correction. Feedback can also be accumulated beyond a
single interaction. On-Policy Data Evolution~
\cite{huang2026onpolicydataevolutionvisualnative} uses failures produced by the
agent's own visual-search trajectories to determine which future trajectories
should receive additional training pressure, while ReTools~
\cite{dong2026retoolsreflectionenhancedtool} combines tool calling with
reflection and revises tool choices when the observed trajectory reveals a
semantic mismatch. These approaches convert observed execution failures into
signals for immediate correction or future policy improvement.

Feedback-adaptive execution therefore closes the loop between analytical intent
and observed tool behavior: tool-requirement signals define what successful
execution should achieve, tool-specification signals constrain how the agent can
respond, and tool-feedback signals determine whether revision is necessary.

\paragraph{Technical Route Summary.}
The three execution routes address complementary problems in translating an
analytical plan into reliable tool-mediated actions over the data environment.
\textbf{Tool Capability Boundary Learning} asks, ``Can this tool reliably
satisfy the current analytical requirement?'' It aligns tool requirements with
declared capabilities and refines this alignment through observed behavior.
\textbf{Compositional Tool Reasoning} asks, ``How should multiple tool-mediated
operations be connected?'' It combines analytical sub-goals, tool dependencies,
and intermediate execution results into a coherent sequence.
\textbf{Feedback-Adaptive Tool Calling} asks, ``Does the observed execution
indicate that the current tool use should be revised?'' It evaluates tool
behavior against the analytical objective and uses execution feedback to change
subsequent calls or improve future tool-use policies.

Across the three routes, the same signal sources serve different functions.
Tool-requirement signals keep execution anchored to the evolving data state and
analytical objective; tool-specification signals connect these requirements to
the capabilities and dependencies exposed by the execution environment; and
tool-feedback signals return evidence from actual interaction with the data.
Together, they form the information loop through which a Data Agent turns an
abstract plan into reliable, data-grounded execution.

\begin{figure}[t]
\centering
\includegraphics[width=0.95\linewidth]{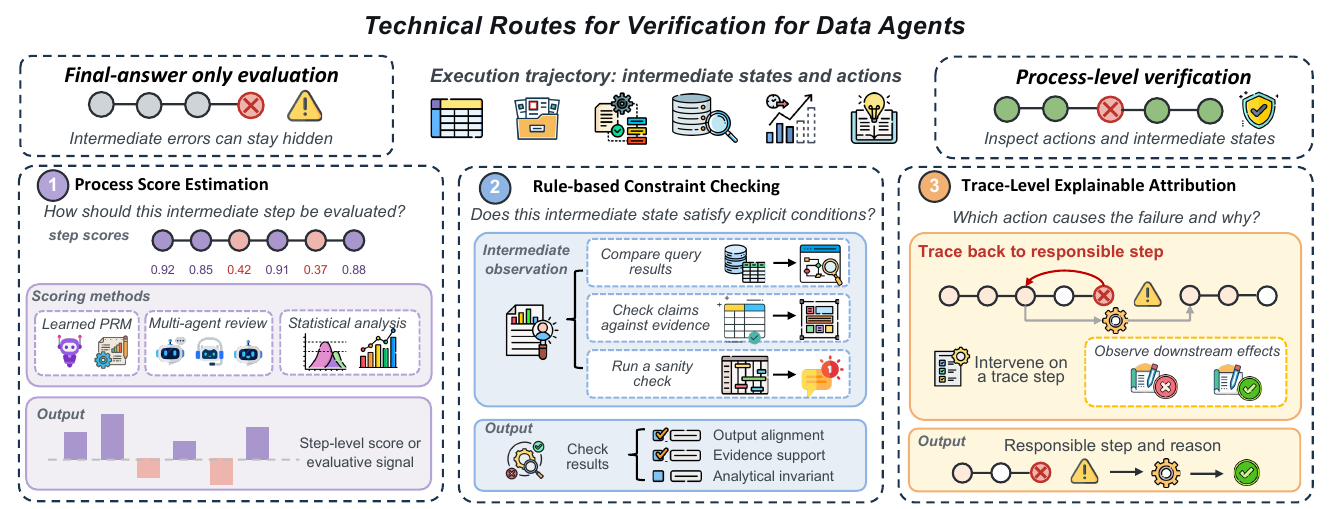}
\caption{Technical routes for verification: Process Score Estimation,
Rule-based Constraint Checking, and Trace-Level Explainable Attribution.}
\label{fig:verification_overview}
\end{figure}


\subsection{Verification}
\label{sec:verification}

For a Data Agent, reliable verification requires evaluating not only whether the final output is correct, but whether intermediate states and actions are logically sound and grounded in evidence. This becomes especially critical in long-horizon tasks, where early errors can easily cascade into incorrect final results and misguide subsequent repairs. The literature therefore suggests three primary verification routes: \textbf{Process Score Estimation}, \textbf{Rule-based Constraint Checking}, and \textbf{Trace-Level Explainable Attribution}. These verification routes are summarized in Figure~\ref{fig:verification_overview}.

\subsubsection{Process Score Estimation}

Process score estimation derives evaluative scores or signals for intermediate states and actions before the final outcome is produced. These scores may be obtained through learned reward models, multi-agent review, etc.

\textbf{Structured Data.}
In structured data environments, verification often relies on deterministic observations: SQL queries either produce results or fail, and intermediate tables can be explicitly inspected. Here, process scores are typically learned through SFT and RL to evaluate intermediate actions before they cascade to the final output. MERIT~\cite{wang2026learningretrieveduallevellongterm} uses a Process Reward Model (PRM) over state-memory pairs to score match quality, making retrieved memories verifiable before they influence subsequent SQL generation. Reward-SQL~\cite{zhang2026rewardsqlboostingtexttosql} introduces a PRM that combines trajectory scoring with entropy-based step weighting, transforming intermediate SQL reasoning steps into dense verification signals during both RL training and inference. Both approaches share a core insight: verification should occur immediately after each action rather than at the pipeline's end.

\textbf{Semi-Structured Data.}
In semi-structured data environments, intermediate states are less predictable, making dense process scores harder to obtain directly. One line of work therefore derives verification signals through multi-agent critique and review. Table-Critic~\cite{Yu2025tablecriticmulti_151} similarly coordinates judge, critic, refiner, and curator agents to assess intermediate tabular reasoning and expose local errors before they propagate. TabTracer takes an execution-grounded approach to table reasoning: after each tool action, it checks the resulting table state and assigns a reflection score, which is backpropagated through an MCTS tree to guide subsequent search~\cite{luo2026tabtracer}. Together, these methods obtain process-level signals through reviewer critique or scored execution traces, without requiring a separately trained reward model.
\textbf{Unstructured Data.}
In unstructured data environments, the lack of explicit structure makes it difficult to define and supervise what constitutes a ``correct intermediate step.'' Consequently, learned process scoring remains comparatively underdeveloped, and alternative estimation mechanisms become more important. UQE~\cite{Dai2024queryengineunstructured_164} uses sampling and variance analysis to estimate the reliability of query execution over unstructured collections, providing a statistical process score without relying on a learned reward model. LongDS-Bench~\cite{xu2026longdsbenchfailurelonghorizonagentic} labels failures related to long-range context memory, making otherwise hidden process failures observable. DSAEval~\cite{sun2026dsaevalevaluatingdatascience} further compares complete agent trajectories across real-world tasks, showing that final scores alone can obscure substantial variation in the underlying analytical process. These benchmarks primarily expose the need for process-level verification, highlighting the difficulty of constructing reliable supervision for intermediate states in open-ended data analysis.

\subsubsection{Rule-based Constraint Checking}

Rule-based constraint checking tests whether intermediate observations satisfy specific constraints. Unlike process score estimation, this route relies less on graded evaluative signals and more on explicit validity conditions.

\textbf{Structured Data.}
In structured data environments, constraints are well-defined through data structures. Verification here relies on strict rule checks applied to candidate actions before execution. For instance, CHASE-SQL and OpenSearch-SQL enforce \textit{logical equivalence rules} via output-alignment checks, discarding candidate SQL queries that violate consistency constraints~\cite{Pourreza2025chasemultipath_83,Xie2025opensearchenhancingtext_215}. Similarly, E2ETune and $\lambda$-Tune validate generated DBMS configurations against strict \textit{performance bounds} using cost models, ensuring these configurations adhere to system safety rules before being trusted~\cite{Huang2025e2etuneknobtuning_112,Giannakouris2025tuneharnessinglarge_111}. These methods leverage the deterministic properties of databases to turn abstract verification into concrete constraint checking.

\textbf{Semi-Structured Data.}
In semi-structured data environments, verification must account for relationships that are implicit in table layouts and analytical workflows. TabClaw checks whether the completed analysis addresses the user's request and whether its conclusions are supported by the observed table evidence~\cite{cheng2026tabclawinteractiveselfevolvingagent}. TaPERA grounds long-form table answers in the outputs of executable programs, allowing generated claims to be checked against computed results~\cite{Zhao2024taperaenhancingfaithfulness_28}. For tables with hierarchical headers and merged cells, ST-Raptor represents the layout as a tree and applies forward validation to its execution steps and backward validation to answer reliability~\cite{Tang2025raptorpoweredsemi_136}. These approaches use different checks, but each makes part of the agent's reasoning or output verifiable against the table and the task.

\textbf{Unstructured Data.}
In unstructured environments, explicit structural constraints are largely unavailable, so verification increasingly relies on explicit semantic or analytical criteria. MuDABench and UniDataBench illustrate this need at the evaluation level by assessing whether intermediate evidence remains sufficiently grounded in the underlying documents~\cite{li2026navigatinglargescaledocumentcollections,weng2026unidatabench}. At the method level, sanity-checking approaches convert common analytical invariants and domain assumptions into lightweight verification rules~\cite{rewolinski2026sanitychecksagenticdata}. These approaches extend constraint checking from deterministic structural rules to explicit semantic and validity conditions.

\subsubsection{Trace-Level Explainable Attribution}

Trace-level explainable attribution identifies which execution step caused an explicit failure and why. Its key challenge is to trace how a state change introduced by one action propagates to downstream failure. When data structures and properties are explicit, these effects are easier to track and isolate; as they become partial or latent, attribution increasingly requires broader trajectory-level analysis.

\textbf{Structured Data.}
In structured data environments, intermediate states and dependencies are
often explicitly represented, making the effect of individual actions easier
to trace through subsequent execution. Attribution can therefore isolate a
suspected step by intervening on the execution trace and measuring how the
downstream outcome changes. For instance, CausalFlow computes a Causal
Responsibility Score by systematically intervening on individual trace steps
and observing their downstream effects, thereby identifying steps that
contribute most strongly to the observed failure
~\cite{bonagiri2026causalflowcausalattributioncounterfactual}. REFLECT similarly preserves verified interventions so that previous repairs can inform later error attribution~\cite{lin2026reflectinterventionsupportederrorattribution}. These methods turn
fault localization into an explicit trace-level analysis rather than merely
associating the terminal failure with the most recent action.

\textbf{Semi-Structured Data.}
In semi-structured data environments, intermediate execution states often provide only partially explicit data properties as correctness anchors. Because these local properties may be insufficient to isolate the exact faulty step, attribution shifts toward analyzing the broader execution trajectory. For instance, FAMA analyzes failure trajectories from baseline agents to identify prevalent error patterns, subsequently routing them to specialized meta-agent strategies based on these observations for targeted verification~\cite{saeidi2026famafailureawaremetaagenticframework}. These approaches provide an explainable basis for diagnosing and correcting complex execution errors.

\textbf{Unstructured Data.}
In unstructured environments, it is difficult to verify a single execution step in isolation due to the lack of explicit data properties. Therefore, attribution is performed by tracing failures backward through the execution history. For example, \textit{Failure is Feedback} models information retrieval as a sequential decision process, utilizing failed retrieval paths as diagnostic signals~\cite{yun2026failurefeedbackhistoryawarebacktracking}. Rather than blindly retrying, it summarizes the terminal failure into actionable feedback to identify which historical actions may have contributed to the dead end, using this insight to guide its backtracking. By explicitly mapping downstream failures back to their historical origins, this approach achieves explainable attribution.

\paragraph{Technical Route Summary.}
These routes address complementary dimensions of the
verification problem. Process score estimation asks, ``How should this
intermediate step be evaluated?'' It derives process scores or evaluative
signals for intermediate states and actions before the final outcome is
produced. Rule-based constraint checking asks, ``Does this intermediate state
satisfy explicit conditions?'' It tests states against predefined validity
constraints. Trace-level explainable attribution asks, ``Which action causes
the failure and why?'' It analyzes the execution trajectory to locate the
responsible step and provide a reason when an incorrect result is produced.

The first two routes determine whether intermediate data states remain
valid through explicit criteria, whereas trace-level attribution
localizes the source of a detected failure. Process score estimation is
predominantly \textit{scoring-driven}: it derives intermediate verification
scores through learned reward models, multi-agent agreement, statistical estimation, etc. Rule-based constraint
checking is predominantly \textit{constraint-driven}: it evaluates intermediate
states against explicit validity conditions. Trace-level explainable attribution
is predominantly \textit{trace-analysis-driven}: it uses failure-pattern
analysis to attribute observed failures to earlier execution steps.

\begin{figure}[t]
\centering
\includegraphics[width=1\linewidth]{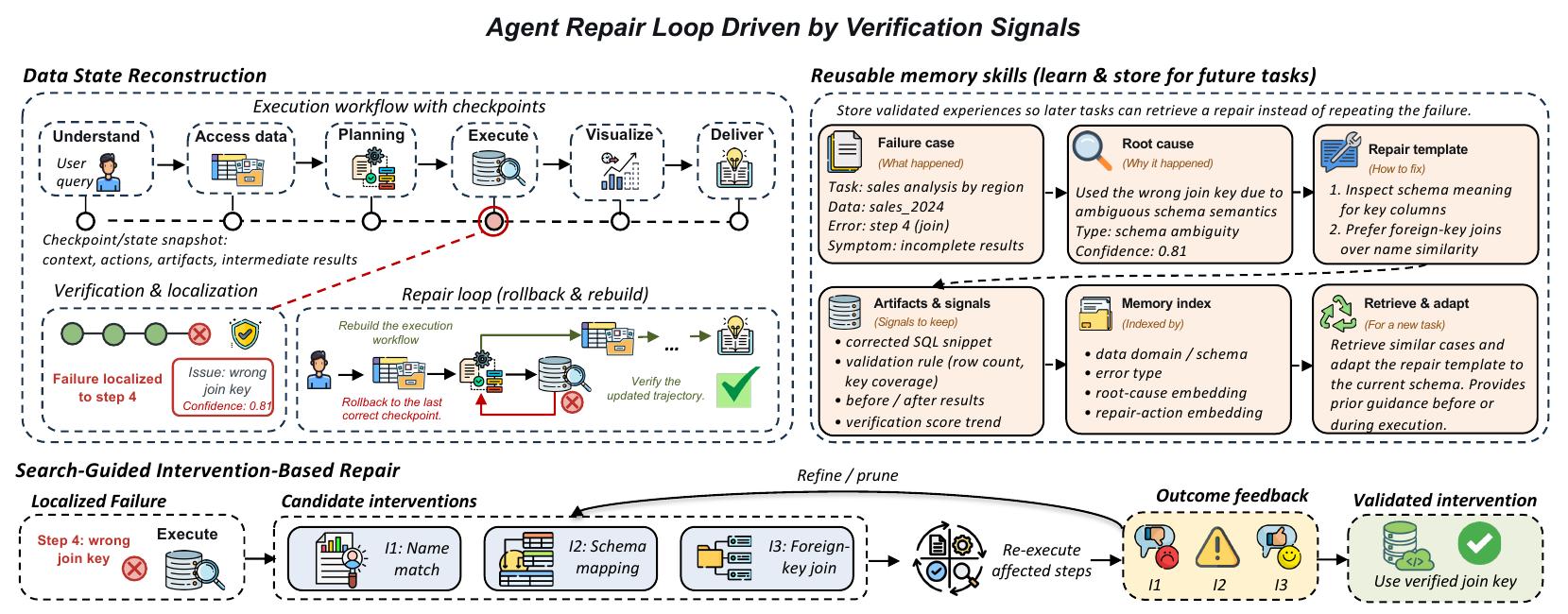}
\caption{Technical routes for repair: Data State Reconstruction,
Search-Guided Intervention-Based Repair, and Reusable Memory Skills.}
\label{fig:agent_repair_combined}
\end{figure}

\subsection{Repair}
\label{sec:repair}

The repair phase leverages verification outcomes to correct the workflow. Its objective is not only to resolve the current failure efficiently, but also, where possible, to retain generalizable lessons for future failures. Accordingly, reliable repair involves three complementary decisions: where recovery should begin, how the detected failure should be corrected, and what should be retained from a validated repair for future reuse. The literature addresses these decisions through three corresponding technical routes: data state reconstruction restores a consistent state from which execution can resume, search-guided intervention-based repair explores and validates candidate corrections, and reusable memory skills retain validated repair experience for subsequent tasks. These repair routes are summarized in Figure~\ref{fig:agent_repair_combined}.

\subsubsection{Data State Reconstruction}

Data state reconstruction recovers a failed workflow by rolling back to a consistent intermediate state rather than restarting from scratch. The core research question is not whether to save states, but how to determine which state is recoverable and what must be recomputed. These decisions dictate whether the repair is safe, minimal, and effective.

\textbf{Structured Data.}
In structured data environments, intermediate states and constraints are often explicitly represented, making selective recovery more tractable. Reconstruction therefore focuses on identifying a valid restore state and recomputing only the affected portion of the workflow. SagaLLM~\cite{chang2025sagallmcontextmanagement} decomposes long-horizon workflows into compensable steps with corresponding recovery actions, allowing failed execution to be rolled back without restarting the entire workflow. DART~\cite{yang2026dartsemanticrecoverability} further introduces semantic recoverability, selecting restore states that remain valid under constraints so that recovery does not invalidate already committed downstream work. These systems aim to restore a consistent data state while minimizing unnecessary recomputation.

\textbf{Semi-Structured Data.}
In semi-structured data environments, repairing an earlier step may change formulas and cell values on which later steps depend. Pista~\cite{sabouri2026auditing} caches the sheet state at each step. When a user corrects an earlier action, it branches from the edited state and regenerates subsequent steps while preserving the original branch for comparison. This supports recovery without discarding unaffected spreadsheet work.

\textbf{Unstructured Data.}
In unstructured data environments, a failed retrieval–reasoning workflow cannot always be restored by replaying a fixed sequence of operations. The agent must determine which retrieved evidence and intermediate reasoning remain valid before continuing. Doctor-RAG~\cite{jiao2026doctorragfailureawarerepairframework} identifies the earliest failure point in the trajectory, preserves the validated reasoning prefix and retrieved evidence, and applies a targeted repair from that point. This allows the workflow to resume from an intermediate state without repeating earlier retrieval and reasoning steps.

\subsubsection{Search-Guided Intervention-Based Repair}

Search-guided intervention-based repair explores repair variants to a suspected faulty step and performs the step again to determine which intervention restores successful execution. Unlike attribution, this route searches over possible repairs and uses subsequent execution outcomes to decide which variant should be retained.

\textbf{Structured Data.}
In structured data environments, explicit data constraints make it easier to locate the subspace of relatively optimal repairs. AgentFixer combines failure detection with root-cause analysis to identify recurrent failure patterns and narrow repair variants to specific prompting or coding changes~\cite{mulian2026agentfixerfailuredetectionfix}. These modifications are then reevaluated through subsequent execution, allowing the system to determine whether the proposed intervention mitigates the observed failure. Here, search is achieved by progressively restricting the repair space rather than exhaustively enumerating possible repairs.

\textbf{Semi-Structured Data.}
In semi-structured data environments, repair search must account for partially explicit data states, so a locally plausible modification may still fail after interacting with later data steps. The Self-Healing Framework addresses this through adaptive replanning and corrective prompting, using detected failures and subsequent execution results to revise the recovery strategy~\cite{jeong2026selfhealingframeworkreliablellmbased}. Repair therefore proceeds as an iterative search over alternative interventions, where unsuccessful modifications trigger further adjustment rather than being accepted immediately.

\textbf{Unstructured Data.}
In unstructured data environments, the repair space becomes broader because modifications to one data step may interact unpredictably with the rest of the workflow. MOAR addresses this problem through global workflow search, using a multi-armed bandit framework to prioritize which rewritten workflows should be explored and evaluating candidate rewrites at the level of the entire workflow~\cite{wei2026multiobjectiveagenticrewritesunstructured}. Rather than repairing each step independently, it searches over interacting workflow modifications and retains rewrites according to their downstream performance. This makes repair a broader search problem when the effects of local interventions cannot be evaluated independently.

\subsubsection{Reusable Memory Skills}

Reusable memory skills extend repair beyond the current failure by retaining
validated corrections for future reuse. Rather than repeatedly rediscovering
the same repair, the agent converts successful repair experience into
retrievable memory or internalized skills that can guide later recovery.
A useful repair experience therefore captures not only what failed, but under
what conditions the correction succeeded and when it may apply again.

\textbf{Structured Data.}
In structured data environments, failure and repair behaviors can often be represented explicitly through data states, making repaired experience easier to abstract into reusable artifacts. MERIT stores successful and failed trajectories in dual-level memory and learns which experience should be retrieved for the current task~\cite{wang2026learningretrieveduallevellongterm}. CausalFlow further converts validated failures into contrastive repair trajectories that can support later recovery or training~\cite{bonagiri2026causalflowcausalattributioncounterfactual}. The common principle is to compress a repaired trajectory into a more reusable representation rather than replaying the entire original execution.

\textbf{Semi-Structured Data.}
In semi-structured data environments, failure and repair behaviors are more dependent on the local data state, making exact repair examples less directly reusable. Existing approaches therefore focus more on extracting recurring repair patterns from multiple failure trajectories and internalizing them as reusable skills. DataCOPE contrasts successful and unsuccessful trajectories to discover transferable skills that can guide later agents~\cite{qiu2026unsupervisedskilldiscoveryagentic}. Fission-GRPO and related reflection-based training methods instead turn execution failures into supervision, allowing recovery behavior to be gradually incorporated into the model itself~\cite{zhang2026robusttoolusefissiongrpo,su2026failure}. In this setting, reuse shifts from retrieving specific past repairs toward learning repair strategies.

\textbf{Unstructured Data.}
In unstructured data environments, useful recovery experience may depend on the evidence retrieved and the reasoning errors encountered in earlier tasks. Rather than storing a fixed repair template, recent systems abstract past successes and failures into guidance that can be recalled when similar conditions arise. FinAcumen~\cite{guo2026finacumen} distills successful strategies and failure-derived cautionary rules from scored financial reasoning trajectories, retrieving them selectively for subsequent tasks. DS-MCM~\cite{sun2026deep} constructs success and failure memories from historical deep-search trajectories and retrieves relevant entries when it detects a mismatch between retrieved evidence and the agent's reasoning, using them to guide corrective intervention. These systems demonstrate cross-task reuse of failure-informed experience.

\paragraph{Technical Route Summary.}
The three repair routes address complementary functions in the repair
lifecycle. Data state reconstruction asks, ``Where should recovery begin?''
It restores a consistent data state from which affected steps can be
recomputed. Search-guided intervention-based repair asks, ``How should the
detected failure be corrected?'' It explores candidate modifications and
validates them through subsequent execution. Reusable memory skills ask,
``What should be retained from a validated repair?'' They convert successful
corrections into retrievable experience or learned skills that can support
future recovery.

Together, these routes connect immediate recovery with longer-term reuse.
Data state reconstruction is \textit{restoration-driven}: it uses data constraints to identify a recoverable state and recompute forward. Search-guided intervention-based repair is \textit{search-driven}: it explores and refines repair variants according to their downstream execution outcomes until a successful repair is found. Reusable memory skills are \textit{experience-driven}: they transform repaired trajectories into reusable representations or learned behaviors that can support later tasks.

\section{Open Reliability Problems and Future Directions}

\begin{figure}[t]
\centering
\includegraphics[width=0.95\linewidth]{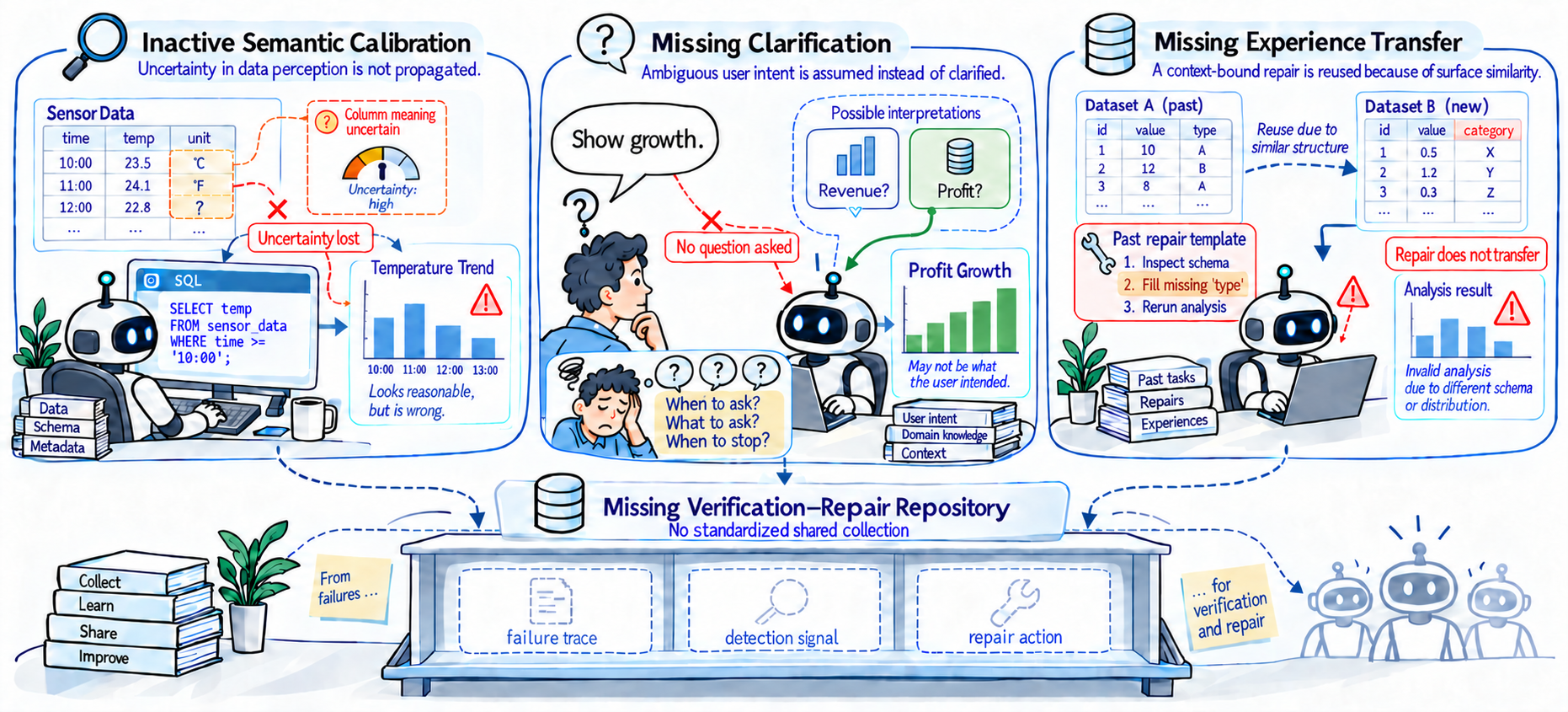}
\caption{We identify four open reliability problems that prevent Data Agents from producing reliable analytical outcomes: (1) the Inactive Semantic Calibration Problem, (2) the Missing Clarification Problem, (3) the Missing Experience Transfer Problem, and (4) the Missing Verification-Repair Repository Problem.}
\label{fig:open_reliability_problems}
\end{figure}

Despite significant progress in Data Agents, several open reliability problems remain unresolved across the workflow harness. Based on the technical routes reviewed in the preceding sections, we identify four such problems concerning how agents calibrate their interpretation of data, handle ambiguity, transfer experience across tasks, and accumulate reusable verification-and-repair knowledge. An overview is provided in Figure~\ref{fig:open_reliability_problems}. These problems span multiple workflow stages and cannot be addressed by improving an individual model or tool in isolation. We discuss each problem below and outline corresponding directions for future research.

\textbf{The Inactive Semantic Calibration Problem.}
This problem arises not from vague instructions, but from how the agent handles uncertainty in its data perception. Even when a user's task intent is perfectly clear, the agent's understanding of the underlying data, such as its exact distributions, implicit schema semantics, or edge cases, is rarely absolute. A column might appear relevant or a sampled distribution might suggest a certain trend, but this perception inherently carries uncertainty. The fundamental flaw in current workflows is the failure to propagate this uncertainty downstream. A confidence score attached to a schema link or a probabilistic assessment of a data distribution does not carry through to SQL generation, table joins, or visualization. Instead, the agent collapses its probabilistic perception into a single deterministic assumption. Because this assumption is syntactically valid, it rarely triggers an exception. The workflow proceeds silently, magnifying the initial miscalibration into a polished but fundamentally flawed chart, report, or model, leaving downstream steps unable to condition on the true reliability of the data state.

\textbf{The Missing Clarification Problem.}
In contrast to semantic calibration, which deals with uncertain data perception, this problem occurs when the user's task intent itself is inherently ambiguous or underspecified. Analytical requests are often vague: terms such as ``growth,'' ``risk,'' or ``regional performance'' admit multiple valid interpretations. Depending on the specific interpretation, the agent can yield entirely different results that are all technically executable. The failure here is not a misinterpretation of the data, but the fact that the agent proceeds as if the task were fully specified, guessing the task's intent rather than halting to ask. Current agents typically default to the most statistically probable interpretation because they lack a disciplined mechanism for deciding when they are authorized to make assumptions versus when they must seek user alignment. Even when clarification questions are generated, there is no rigorous standard for what constitutes a useful analytical question, nor a stopping criterion: agents either ask too little and proceed with hallucinated assumptions, or ask too much and induce user fatigue.

\textbf{The Missing Experience Transfer Problem.} 
This problem highlights the fragility of how Data Agents internalize post-failure corrections. While many agents can store a successful trajectory, retrieve a past repair, or learn recovery behavior through training, current transfer mechanisms still rely heavily on task similarity. They perform well when a new task shares similar data structures or statistical distributions with the original failure. However, their effectiveness often degrades as the new task departs from these conditions. Experience reuse can therefore remain sensitive to surface similarities rather than fully capturing which aspects of a repair are broadly reusable. A sequence of steps that precedes a successful outcome may contain both generally useful repairs and those that are incidental to that specific dataset. This can lead to negative transfer: when faced with tasks involving new data structures or shifted statistical distributions, the agent may retrieve or apply a multi-step repair template based on superficial similarities even though some of its underlying assumptions no longer hold. As a result, the applied repair may become mismatched to the new data environment, causing the workflow to fail or produce invalid results.

\textbf{The Missing Verification-Repair Repository Problem.}
Compounding the difficulty of experience transfer is the lack of large-scale, standardized collections of failures and repairs in machine learning and data analysis. Existing systems typically accumulate verification and repair experience within individual tasks or frameworks, while the community lacks shared repositories that systematically capture faulty analytical trajectories, the signals used to detect failures, and repair actions. As a result, verification mechanisms have limited failure evidence from which to learn, while repair agents have limited reusable experience for determining how detected failures should be corrected. Each system must therefore construct its own small and often narrow collection of failure cases for training or evaluation. This limitation makes it difficult to train agents to recognize diverse failure modes, verify intermediate analytical states, and generalize repair strategies beyond previously observed datasets. Building such resources is particularly challenging because data-science failures are highly heterogeneous and wide-ranging, ranging from data-formatting anomalies and silent distribution shifts to inappropriate statistical assumptions and algorithmic misconfigurations. A shared failure-and-repair repository could therefore provide reusable experience for repairing failures, while serving as a testbed for evaluating verification and repair.


\section{Horizontal Analytical Task Taxonomy}

This section presents a task-oriented taxonomy of autonomous Data Agents by organizing analytical tasks according to the objectives they are intended to accomplish. This perspective differs from the workflow taxonomy in the previous section, which describes \emph{how} Data Agents operate through perception, planning, execution, verification, and repair. Here, we instead focus on \emph{what} analytical objectives agents are expected to accomplish. The distinction is important because a single analytical objective may require multiple workflow stages, while the same workflow operation may support very different analytical objectives. For example, answering a question over a database may require schema inspection, query generation, execution, and result interpretation, whereas exploring an unfamiliar dataset may involve many of the same operations but pursue a fundamentally different analytical goal.

We organize these objectives horizontally across several task families that can arise in different application domains. These categories are not intended to prescribe a fixed execution order. In practice, an agent may move between them iteratively, and a single analytical task may involve multiple task families. For instance, an analysis may require collecting external data, preparing heterogeneous sources, querying relevant records, and finally synthesizing evidence.

This task-oriented view also distinguishes analytical objectives from individual data-processing operations. Operations such as missing-value imputation, feature transformation, schema matching, table joining, retrieval, and code execution may serve as intermediate actions within a broader task. They become task-level objectives when the preparation, retrieval, or transformation itself constitutes the goal of the analysis. This distinction allows the taxonomy to remain orthogonal to both the workflow organization and the application-oriented taxonomy presented later, where different task capabilities are combined within specific Data Agent settings.

\subsection{Data Collection and Acquisition}

Data collection and acquisition concerns obtaining the data or information required for a downstream analytical objective~\cite{ma2026autodata,hu2022multi}. Unlike tasks that assume a given dataset, agents in this setting must determine what information is needed, identify appropriate sources, and acquire relevant data through available interfaces. Depending on the setting, this may involve dataset discovery, web-based information gathering, database access, or the retrieval of external resources~\cite{lomborg2014using}.

The agent therefore needs to connect a high-level analytical objective with concrete data sources. Rather than treating retrieval as a single lookup operation, autonomous systems may need to formulate search queries, inspect candidate sources, determine whether the collected information is relevant, and iteratively acquire additional data when the initial evidence is insufficient~\cite{fu2025autonomous}. This makes data collection closely coupled with planning and source selection, particularly when the required information is distributed across multiple sources or is not known in advance.

\subsection{Data Preparation and Transformation}

Data preparation and transformation concerns converting collected or existing data into a form suitable for the intended analysis. Representative tasks include identifying problematic values, cleaning inconsistent records, transforming data according to natural-language instructions, and integrating information across different representations. These tasks can range from individual repair operations to longer sequences in which an agent must inspect the data, determine appropriate transformations, and execute them in combination.

Unlike fixed preprocessing pipelines, autonomous Data Agents must often infer which preparation operations are appropriate from the data and the analytical objective. An agent may need to inspect distributions or metadata, identify anomalous values, determine whether an apparent irregularity is actually an error, and select an appropriate transformation. In more complex settings, preparation becomes iterative: the outcome of one operation can affect the choice of subsequent operations, requiring the agent to inspect intermediate results and revise its actions.

The preparation process consequently serves not only to improve data quality but also to establish a suitable representation for downstream reasoning. Systems that address context-aware cleaning consider the relationship between detected data problems and their surrounding context~\cite{Biester2024llmcleancontextaware_42}. Retrieval-based approaches use relevant examples or prior knowledge to guide cleaning decisions~\cite{Naeem2024retcleanretrievalbased_48}, while evolutionary approaches explore how language-model-based strategies can be refined through iterative feedback~\cite{Gong2025evolutionarylargelanguage_44}. More recent work frames data preparation as an agentic process in which multiple operations are coordinated to prepare data for subsequent analysis~\cite{wang2026dataforgeagenticplatformautonomous,xu2026prepbenchfarnaturallanguagedrivendata}. These settings illustrate that preparation quality cannot be reduced to whether individual transformations execute successfully: the agent must also preserve the semantics of the data and produce a representation that remains appropriate for the intended analytical task.

\subsection{Data Querying and Information Seeking}

Data querying and information seeking concern answering well-defined analytical questions by locating and operating on relevant data. In structured settings, this includes \textbf{Text-to-SQL}, \textbf{Table QA}, and \textbf{Tabular Reasoning}, where an agent maps natural-language requests to executable queries or programs that retrieve, aggregate, compare, and reason over table contents. Related tasks can also operate over graph-structured databases or other queryable data sources.

LLM-based agents typically ground the user's request in schemas, values, and available data structures before generating executable operations. PV-SQL, for example, incorporates database probing to obtain additional schema and value information before query generation~\cite{tian2026pvsqlsynergizingdatabaseprobing}. FlexSQL explores candidate queries and uses execution feedback to iteratively refine the generated query~\cite{pham2026flexsql}. Other approaches employ multi-path generation and execution-based verification to improve complex text-to-SQL reasoning~\cite{Pourreza2025chasemultipath_83,Xie2025opensearchenhancingtext_215}. The same analytical objective can extend beyond conventional relational tables: recent work considers multimodal tables and multi-agent reasoning, where tabular information is analyzed together with other evidence sources~\cite{kwok2026tabqaworldoptimizingmultimodalreasoning,Yu2025tablecriticmulti_151}.

The defining challenge is therefore not simply producing an executable query, but aligning the operation with what the user actually intends to retrieve or compute. Query execution provides a relatively direct way to verify whether the produced operation returns the expected result, yet a syntactically valid and executable query can still select the wrong column, value, or relationship. Data querying consequently provides a useful setting for studying the gap between executable correctness and semantic correctness, particularly when natural-language requests must be grounded in complex or ambiguous data structures.

\subsection{Data Analysis and Prediction}

Beyond querying existing data, Data Agents may be required to analyze data and derive predictions or decisions from observed patterns. This task family covers analytical objectives in which the agent must identify relevant patterns, select appropriate analytical methods, and produce predictions, classifications, or diagnostic results. Representative subtasks include \textbf{Time Series Forecasting}, \textbf{Time Series Classification}, and \textbf{Anomaly Detection}, where temporal structure provides an important source of information but is not itself the defining task objective.

LLM-based agents typically delegate numerical modeling to specialized statistical or machine learning tools rather than performing the analysis entirely within the language model. The agent may first inspect the data, assess its quality and structure, select an appropriate model or analytical procedure, and then interpret the resulting outputs. Systems such as TSQAgent and TimeClaw illustrate this pattern by assessing time series quality and identifying characteristics such as periodicity and anomalies to guide subsequent analytical actions~\cite{wu2026tsqagentratingtimeseries,liu2026timeclawtimeseriesaiagent}. Tool-augmented sequential reasoning and long-horizon scientific agents further demonstrate how analytical modeling can be embedded within planning, verification, and repair processes~\cite{tao2026castr1learningtoolaugmentedsequential,zhan2026aionnextgenerationtaskspractical}. The central challenge is therefore not only to obtain a numerically valid result, but also to select analytical assumptions and methods that are appropriate for the data and the intended objective.

\subsection{Data Interpretation and Multimodal Reasoning}

Some analytical objectives cannot be reduced to answering a predefined query or producing a single prediction. Instead, the agent must explore data, identify potentially useful patterns, and synthesize evidence from sources that may differ in structure or modality. This task family covers \textbf{Exploratory Data Analysis}, \textbf{Visualization and Insight Generation}, and \textbf{Document and Multimodal Reasoning}. For exploratory analysis, agents may profile datasets, generate visualizations, inspect intermediate results, and iteratively refine their analysis. Text-to-visualization systems and visualization agents translate analytical intentions into chart specifications or executable plotting programs, while systems such as MatPlotAgent use generated figures as feedback for subsequent refinement~\cite{xu2026reliableagenticprogressivetexttovisualization,Yang2024matplotagentmethodevaluation_152,Chen2025interchatenhancinggenerative_130,Islam2024datanarrativeautomateddata_154}. Interactive BI systems further connect exploration with conversational context and iterative analytical interaction~\cite{li2026twinbiagenticdigitaltwin,Jiang2025siriusbicomprehensivepowered_159,Zhao2024chat2datainteractivedata_225}. When evidence is distributed across unstructured or multimodal sources~\cite{samuelsen2026mimirragmultiagentragframework,li2026navigatinglargescaledocumentcollections,Shankar2025docetlagenticquery_198}, the same objective requires additional retrieval and grounding capabilities. Document and multimodal systems combine retrieval, parsing, visual grounding, and tool use to locate and interpret relevant evidence across long documents, webpages, charts, and video~\cite{SaadFalcon2024pdftriagequestionanswering_85,guo2025seeingevaluatingmultimodalllms,zhang2026navigatingmiragedualpathagentic,liu2026retoolvideorecursivetoolusingvideo}. These tasks highlight a broader requirement for autonomous Data Agents: useful analysis depends not only on generating an answer, but also on identifying relevant evidence and producing findings that remain faithful to the underlying data.

\section{Vertical Application Settings and Task Composition}

The analytical task taxonomy presented in the previous section characterizes the capabilities that autonomous Data Agents can provide, ranging from data acquisition and preparation to querying, analysis, prediction, visualization, and multimodal reasoning. In practical settings, however, these capabilities are rarely deployed independently. Real-world analytical applications typically compose multiple capabilities within a larger workflow, with their composition shaped by the degree of user interaction, collaboration, autonomy, and domain knowledge required~\cite{feng2026graph,xiang2026systematic}. For example, an interactive analytical assistant may combine natural language querying, data transformation, analysis, and visualization in response to successive user requests, whereas an autonomous data workflow may coordinate data preparation, database operations, analytical modeling, and runtime verification without continuous human intervention.

This section therefore examines how these horizontal capabilities are composed across vertical application settings. Rather than introducing additional analytical task categories, we consider how the capabilities identified in the previous section are combined within different modes of analytical work. The same analytical capability may appear across multiple settings, but its role and composition change according to the surrounding workflow and operational requirements.

\subsection{Interactive Data Assistance}

Interactive Data Assistance represents applications in which Data Agents work closely with users through natural-language interaction, progressively translating analytical intentions into executable operations and interpretable results. Conversational business intelligence is a representative setting, where agents support users in querying enterprise data, inspecting analytical results, and iteratively refining analytical questions through dialogue.

Commercial systems increasingly reflect this interactive analytical paradigm. \textbf{Fabi Analyst Agent}~\footnote{\url{https://www.fabi.ai/product/analyst-agent}} connects directly to organizational data sources and supports natural-language analysis, Python- and SQL-based computation, dashboard generation, and collaborative access control, with the agent also available through Slack. V7's \textbf{AI Spreadsheet Analysis Agent}~\footnote{\url{https://www.v7labs.com/agents/ai-spreadsheet-analysis-agent}} allows users to upload complex spreadsheets and ask analytical questions in natural language, supporting automated calculations, summaries, and visualizations. These systems illustrate how interactive Data Agents are moving beyond conversational query answering toward persistent analytical environments in which users can iteratively ask questions, inspect results, and refine analyses within the same workspace. The defining characteristic of this application setting is therefore not a particular data modality, but the integration of multiple capabilities around an evolving user-defined objective.

\subsection{Collaborative Analytical Work}

A second application setting involves analytical work performed through collaboration among multiple agents or between agents and specialized analytical components. Here, task composition becomes explicit because different parts of an analytical workflow may be assigned to different agents, tools, or reasoning modules. Instead of requiring a single agent to perform every operation, collaborative systems can distribute analytical responsibilities across specialized components while coordinating their intermediate results.

Commercial tools increasingly support this form of analytical collaboration within existing work environments. \textbf{GPT for Work}~\footnote{\url{https://gptforwork.com/}} integrates AI agents directly into Microsoft Excel and Google Sheets, where they can generate and repair formulas, clean and reorganize data, create charts and pivot tables, and process spreadsheet rows in bulk. \textbf{Codex}~\footnote{\url{https://openai.com/codex/}} provides a related workflow for data science and analytics teams, supporting the integration of data, code, and analytical artifacts within a shared working process. These systems illustrate how agents can take responsibility for specific analytical operations while remaining embedded in a broader human-guided workflow. The resulting application pattern is a division of analytical labor in which users, agents, and specialized tools contribute complementary capabilities and coordinate through shared analytical results.

\subsection{Autonomous Data Workflows}

Autonomous Data Workflows move beyond interactive assistance toward applications in which agents execute a sequence of dependent analytical operations with limited continuous supervision. Such workflows may begin with data acquisition or profiling, proceed through preparation and transformation, invoke analytical or database tools, and iteratively adjust subsequent actions according to execution feedback. The defining feature is therefore not a particular task, but the integration of multiple capabilities where intermediate results influence subsequent operations.

Data preparation and database operations provide natural entry points for this form of autonomy. In a typical workflow, an agent may inspect the available data, determine the required transformations, execute code or queries, evaluate the resulting state, and continue with downstream analysis. This pattern increasingly extends beyond individual data operations toward workflows that combine data engineering, modeling, and system interaction within the same environment.

\textbf{Snowflake CoCo}~\footnote{\url{https://www.snowflake.com/en/product/features/cortex/}} provides a commercial example of this workflow-oriented design. Integrated into the Snowflake platform, CoCo supports data engineering, analytics, machine learning, and agent-building tasks, with an autonomous agent mode that can plan and execute multi-step operations within the user's Snowflake environment. Snowflake's \textbf{Cortex Agents}~\footnote{\url{https://docs.snowflake.com/en/user-guide/snowflake-cortex/cortex-agents}} further allow agents to combine structured data access, unstructured retrieval, and code execution within a single governed workflow, repeatedly selecting tools and evaluating intermediate results before producing an answer. These capabilities illustrate how autonomous analytical workflows can combine database operations while preserving the governance of the underlying data environment.

A similar direction is emerging in integrated data science platforms. \textbf{Databricks Genie Code}~\footnote{\url{https://www.databricks.com/product/genie/code}}
 can automate multi-step data science workflows by planning solutions, retrieving relevant data assets, generating and executing code, inspecting outputs, and fixing errors based on intermediate results. This workflow connects data exploration, coding, execution, and error recovery rather than treating code generation as an isolated operation. The agent can therefore remain within the analytical environment while progressively adapting its actions to the state of the computation.

General-purpose agent platforms extend the same idea to more heterogeneous data workflows. Manus can profile and clean spreadsheets or CSV files, generate charts and reports, and turn completed analyses into reusable workflows for subsequent execution. This supports a transition from one-time analysis toward repeatable data workflows in which similar operations can be applied to updated inputs with less manual intervention. Together, these applications demonstrate how Data Agents are evolving from assistants that perform individual analytical operations into systems that can coordinate preparation, computation, verification, and artifact generation.

\subsection{Knowledge-Intensive Discovery}

Knowledge-Intensive Discovery represents applications in which analytical workflows combine information seeking, evidence synthesis, and domain-specific reasoning with data analysis. In these settings, the agent typically cannot rely on a single structured dataset or a fixed analytical procedure. Instead, it gathers information from external sources, connects heterogeneous evidence, performs quantitative or qualitative analysis, and synthesizes the results into a domain-specific conclusion. This setting is particularly relevant to research, market intelligence, and business analysis, where useful evidence is distributed across documents, webpages, databases, and structured data. \textbf{Manus}~\footnote{\url{https://manus.im/solutions/product}}, for example, supports multi-step research by searching external sources, organizing relevant information, and combining retrieved evidence with structured analysis to produce reports and other analytical artifacts. \textbf{Pelayar's Spreadsheet Agent}~\footnote{\url{https://pelayar.ai/}} provides a more data-centric example by extracting information from PDFs, invoices, images, and other documents and converting the results into structured spreadsheets for subsequent calculation, visualization, and analysis.

Financial and business analysis further illustrates this composition, where agents may retrieve information from financial reports, company disclosures, market sources, and other external documents, interpret textual and tabular evidence, and combine the findings with quantitative analysis. Similar workflows arise when evidence spans multiple modalities, requiring agents to locate relevant information in long documents, extract values from tables or images, interpret charts, and connect these observations with textual or structured data. In such settings, retrieval, multimodal interpretation, analysis, and evidence synthesis are not independent operations. Instead, newly retrieved evidence can change the analytical direction, while intermediate findings can motivate additional information seeking. This iterative interaction between information gathering and analysis distinguishes knowledge-intensive discovery from conventional data analysis and enables Data Agents to support analytical tasks whose conclusions depend on integrating external evidence.

\section{Benchmarking Autonomous Data Agents}

Benchmarking autonomous Data Agents requires evaluating both what analytical capabilities they support and how reliably they execute the underlying analytical process. Existing benchmarks vary in the objectives they assign to agents, the complexity of the data and execution environment, and the criteria used to determine success. We therefore organize the benchmark landscape around these evaluation dimensions, while relating them to the task capabilities and workflow stages introduced above. Representative benchmarks are listed in Table~\ref{tab:data_agent_benchmarks}. This perspective highlights how evaluation is gradually shifting from answer correctness toward the quality, reliability, and efficiency of the overall analytical process.

\begin{table*}[t]
\centering
\caption{Representative benchmarks for evaluating autonomous Data Agents across analytical tasks.}
\label{tab:data_agent_benchmarks}
\tiny
\renewcommand{\arraystretch}{1.15}

\begin{tabular}{p{0.12\textwidth}
                p{0.27\textwidth}
                p{0.27\textwidth}
                p{0.28\textwidth}}
\toprule

\textbf{Benchmark}
& \textbf{Focus}
& \textbf{Task \& Process Coverage}
& \textbf{Evaluation}
\\
\midrule

StockGQL~\cite{liang2024natnl2gqlnovelmultiagentframework}
& Natural-language-to-GQL translation over structured financial knowledge.
& Data Querying; Execution, Verification
& Query correctness and retrieval of the required information.
\\

TableBench~\cite{wu2025tablebench}
& Table question answering covering fact checking, numerical reasoning, data analysis, and visualization.
& Data Querying; Verification
& TableQA accuracy across multiple reasoning categories.
\\

Visual-TableQA~\cite{lompo2026visualtableqaopendomainbenchmarkreasoning}
& Visual reasoning over rendered tables, including structure understanding and multi-step reasoning.
& Visualization and Multimodal Analysis; Verification
& Question-answering and reasoning accuracy on table images.
\\

TopBench~\cite{ji2026topbenchbenchmarkimplicitprediction}
& Implicit predictive reasoning over tabular data, including prediction, decision making, and treatment-effect analysis.
& Analysis and Prediction; Verification
& Performance on analytical objectives beyond direct table lookup.
\\

PrepBench~\cite{xu2026prepbenchfarnaturallanguagedrivendata}
& Natural-language-driven data preparation involving cleaning, restructuring, and table transformation.
& Data Preparation; Execution, Verification
& Correctness of generated output tables across preparation settings.
\\

InfiAgent-DABench~\cite{hu2024infiagentdabenchevaluatingagentsdata}
& End-to-end data analysis over CSV files requiring agents to interact with an execution environment.
& Analysis and Prediction; Planning, Execution
& Automatically evaluated answers across diverse analytical questions.
\\

LongDA~\cite{li2026longdabenchmarkingllmagents}
& Documentation-intensive data analysis requiring retrieval from long documents before computation and code execution.
& Information Seeking, Analysis; Planning, Execution
& Answer accuracy, token efficiency, runtime, and tool interactions.
\\

IDA-Bench~\cite{li2025idabenchevaluatingllmsinteractive}
& Interactive, multi-round data analysis derived from Kaggle notebooks with sequential instructions.
& Analysis and Prediction; Planning, Execution, Verification
& Submission success, baseline achievement, turns, runtime, and generated code.
\\

TableAgent-Bench~\cite{huang2026how}
& Multi-turn table analysis over real-world industrial spreadsheets requiring iterative reasoning and tool use.
& Data Querying; Planning, Execution, Verification
& Task completion across multi-turn table-analysis scenarios.
\\

DataGovBench~\cite{hasegawa2026dataanalysiswild}
& Data analysis over government open data, covering table QA and exploratory insight generation.
& Querying, Visualization; Execution, Verification
& Analytical answers and quality of generated insights.
\\

LongDS-Bench~\cite{xu2026longdsbenchfailurelonghorizonagentic}
& Long-horizon, multi-turn data science tasks with evolving analytical states and intermediate results.
& Analysis and Prediction; Planning, Execution, Verification
& Turn-level accuracy and completion of extended analytical trajectories.
\\

DAComp~\cite{lei2026dacomp}
& Data engineering and open-ended data analysis spanning the data-intelligence lifecycle.
& Cross-task; Planning, Execution, Verification
& Execution-based metrics for engineering and rubric-based analysis evaluation.
\\

FDABench~\cite{wang2026fdabenchbenchmarkdataagents}
& Data analysis over heterogeneous structured, unstructured, and multimodal sources.
& Cross-task; Planning, Execution, Verification
& Answer correctness, report quality, reasoning traces, latency, and token usage.
\\

CODA-BENCH~\cite{zhang2026codabench}
& Data-intensive code-agent tasks requiring data discovery, code generation, and execution.
& Analysis and Prediction; Planning, Execution, Verification
& Data discovery and successful completion of executable analytical tasks.
\\

DataSciBench~\cite{zhang2025datascibenchllmagentbenchmark}
& Data science tasks requiring agents to generate and execute analytical programs.
& Analysis and Prediction; Execution, Verification
& Programmatic evaluation of generated programs and execution results.
\\

AgentGym~\cite{xi2024agentgymevolvinglargelanguage}
& Real-world agent tasks involving tool discovery, selection, and multi-step interaction.
& Cross-task; Planning, Execution, Verification
& Task success and tool-use behavior across interactive environments.
\\

FinRpt~\cite{jin2026finrpt}
& Equity research report generation integrating multiple financial data types.
& Analysis and Prediction; Planning, Execution, Verification
& Multi-dimensional evaluation of generated research reports.
\\

PolitNuggets~\cite{zhu2026politnuggetsbenchmarkingagenticdiscovery}
& Agentic discovery and synthesis of long-tail facts from dispersed information sources.
& Information Seeking; Planning, Execution, Verification
& Evidence discovery, fine-grained factual accuracy, and efficiency.
\\

\bottomrule
\end{tabular}
\end{table*}

\subsection{Benchmark Objectives and Task Coverage}

Existing benchmarks range from focused analytical questions to multi-step and end-to-end data analysis. Structured-data benchmarks typically evaluate whether an agent can answer a question through executable operations, such as generating SQL over relational or graph-structured data~\cite{Pourreza2025chasemultipath_83,Xie2025opensearchenhancingtext_215,Zhou2024nl2gqlmodelcoordination_37}. Other benchmarks target data preparation, asking agents to identify and repair problematic data or execute natural-language transformations~\cite{Biester2024llmcleancontextaware_42,xu2026prepbenchfarnaturallanguagedrivendata,wang2026dataforgeagenticplatformautonomous}. Document and multimodal benchmarks extend the objective to information seeking, evidence synthesis, and visual reasoning~\cite{li2026navigatinglargescaledocumentcollections,guo2025seeingevaluatingmultimodalllms}. More recent benchmarks evaluate complete data-analysis workflows in which agents must coordinate multiple analytical operations over an extended interaction~\cite{lei2026dacomp,sun2026dsaevalevaluatingdatascience,xu2026longdsbenchfailurelonghorizonagentic}.

This progression reflects an important change in benchmark objectives. Rather than evaluating whether an agent can perform one operation correctly, increasingly challenging settings require it to combine data understanding, planning, tool use, and result interpretation within a single task. The distinction is important because success on an individual operation does not necessarily imply that an agent can complete a coherent analytical workflow.

\subsection{Data, Environment, and Task Complexity}

Benchmark difficulty depends not only on the analytical objective but also on the conditions under which the agent must achieve it. Existing evaluations vary along several dimensions:

\begin{itemize}
    \item \textbf{Data complexity:} from clean relational tables to heterogeneous documents, charts, images, video, and multimodal collections~\cite{kwok2026tabqaworldoptimizingmultimodalreasoning,Yu2025tablecriticmulti_151,liu2026retoolvideorecursivetoolusingvideo}.
    \item \textbf{Knowledge requirement:} from self-contained datasets to settings requiring retrieval from large or domain-specific collections~\cite{li2026longdabenchmarkingllmagents,shu2026agenticretrievalaugmentedgenerationfinancial,qian2026deepxivsdkagenticdatainterface,lan2026agenticscholar}.
    \item \textbf{Interaction complexity:} from single-step execution to iterative tool use, persistent state, and long-horizon workflows~\cite{zhan2026aionnextgenerationtaskspractical,chen20266gagentgymtoolusedata}.
    \item \textbf{Environmental uncertainty:} from fixed execution settings to tasks involving unexpected outputs, changing intermediate states, or potential execution failures~\cite{rewolinski2026sanitychecksagenticdata,bertran2026many}.
\end{itemize}

These dimensions determine how much autonomy is required from the agent. In particular, as data heterogeneity and interaction length increase, it becomes harder to separate analytical reasoning from retrieval, tool use, and state management.

\subsection{Evaluation Criteria}

Benchmark evaluation traditionally emphasizes the correctness of the final output, but different analytical settings require more specific criteria. For structured querying, execution accuracy measures whether a generated query produces the expected result~\cite{Pourreza2025chasemultipath_83}, while other work examines faithfulness to the user's intended query~\cite{Zhao2024taperaenhancingfaithfulness_28}. Data preparation can be evaluated through transformation correctness and the quality of the resulting data, while retrieval-oriented tasks additionally require assessment of evidence relevance and sufficiency~\cite{Naeem2024retcleanretrievalbased_48}. Multimodal benchmarks may evaluate visual grounding or whether the correct region, frame, or event is identified~\cite{zhang2026navigatingmiragedualpathagentic,zhang2026rewatch}.

For agentic workflows, evaluation increasingly extends beyond final-output correctness to the quality of the analytical process itself. In addition to answer accuracy, execution accuracy, and overall task success, benchmarks may examine whether the agent selects appropriate tools, produces valid intermediate results, and maintains a coherent trajectory throughout the analysis~\cite{zhang2025deepanalyzeagenticlargelanguage}. Evidence quality is another important dimension, particularly for retrieval- and multimodal tasks, where the relevance, grounding, and faithfulness of the evidence supporting a conclusion can determine whether a correct-looking answer is actually reliable. Evaluation may also consider efficiency and reliability through measures such as the number of execution steps and tool calls, consistency across repeated runs, and robustness to failures~\cite{rewolinski2026sanitychecksagenticdata}. More recent work further examines failure attribution, asking whether an unsuccessful outcome can be traced to a specific intermediate decision or intervention~\cite{lin2026reflectinterventionsupportederrorattribution}.

\subsection{Limitations of Current Benchmarking}

Despite the increasing diversity of benchmark settings, current evaluations still have several limitations in assessing the reliability of autonomous Data Agents. First, final-task correctness does not always reflect whether the agent has performed the intended analysis. An agent may execute a syntactically valid query yet misunderstand the intended schema element or semantic requirement~\cite{Zhao2024taperaenhancingfaithfulness_28}. Similarly, retrieving topically relevant documents does not necessarily mean that the agent has obtained sufficient evidence to support its conclusion~\cite{li2026navigatinglargescaledocumentcollections}. In multimodal settings, an agent may also produce a plausible answer while grounding its reasoning in the wrong visual evidence~\cite{guo2025seeingevaluatingmultimodalllms}. These cases indicate that outcome-based metrics can overlook important failures in semantic alignment, evidence selection, and grounding.

A second limitation is the difficulty of evaluating the analytical process itself. Long-horizon tasks involve multiple dependent decisions, and an error introduced during early planning or tool use may only become apparent several steps later~\cite{xu2026longdsbenchfailurelonghorizonagentic}. Existing benchmarks provide increasing coverage of such workflows, but process quality, efficiency, robustness, and failure recovery remain more difficult to measure consistently than final-answer accuracy. In particular, repeated successful outcomes do not necessarily establish that an agent follows a reliable strategy, while a failed trajectory does not by itself reveal which intermediate decision caused the failure. More comprehensive benchmarking should connect final outcomes with evidence quality, intermediate execution, efficiency, and failure attribution to assess whether agents perform data analysis reliably rather than merely solve individual instances.

\section{Positioning with Existing Surveys}

Existing surveys have organized the emerging field of Data Agents from several
complementary perspectives, including autonomy, analytical capabilities, data
modalities, lifecycle stages, and system architectures. Table~\ref{tab:survey_comparison}
positions this survey against representative efforts by comparing not only
their topical coverage, but also whether they systematically analyze technical
methods and reliability along the agent's operational workflow. Several surveys characterize Data Agents primarily through their degree of
autonomy or the broader interaction between LLMs and data systems
~\cite{zhu2026surveydataagentsemerging,fu2025autonomousdataagentsnew,
zhou2025surveyllmtimesdata}. These perspectives are useful for understanding
how agent responsibilities expand and how increasing autonomy changes system
requirements. However, autonomy level alone does not explain how an agent
handles the distinct operational problems encountered during an analytical
trajectory, such as interpreting heterogeneous data states, selecting and
executing tools, validating intermediate results, and recovering from failures.

\begin{table*}[t]
\centering
\caption{Comparison on Data Agents and related LLM-based data systems across capabilities, data environments, lifecycle
organization, evaluation, stage-wise technical methods, and workflow
reliability. \checkmark~indicates systematic coverage, $\circ$ indicates
partial or secondary coverage, and -- indicates no systematic coverage.}
\label{tab:survey_comparison}

{\scriptsize
\setlength{\tabcolsep}{5.5pt}
\renewcommand{\arraystretch}{1.10}

\begin{tabular}{lcccccc}
\toprule
\textbf{Survey / Study}
& \textbf{Capability}
& \textbf{Data}
& \textbf{Lifecycle}
& \textbf{Evaluation}
& \textbf{Stage-wise}
& \textbf{Reliability} \\
\midrule

Survey of Data Agents~\cite{zhu2026surveydataagentsemerging}
& \checkmark & \checkmark & $\circ$ & $\circ$ & -- & $\circ$ \\

Autonomous Data Agents~\cite{fu2025autonomousdataagentsnew}
& \checkmark & $\circ$ & \checkmark & $\circ$ & -- & $\circ$ \\

LLM $\times$ DATA~\cite{zhou2025surveyllmtimesdata}
& \checkmark & \checkmark & $\circ$ & $\circ$ & -- & -- \\

LLM/Agent-as-Data-Analyst~\cite{tang2025llmagentasdataanalystsurvey}
& \checkmark & \checkmark & $\circ$ & $\circ$ & -- & $\circ$ \\

LLM-Based Data Science Agents~\cite{rahman2025llmbaseddatascienceagents}
& \checkmark & \checkmark & \checkmark & \checkmark & $\circ$ & $\circ$ \\

Clean Up Your Mess~\cite{zhou2026llmscleanmesssurvey}
& \checkmark & \checkmark & $\circ$ & \checkmark & -- & $\circ$ \\

Reasoning and Agentic Systems~\cite{chang2026surveyreasoningagenticsystems}
& \checkmark & -- & \checkmark & \checkmark & $\circ$ & $\circ$ \\

Data Agents: Levels, State of the Art~\cite{luo2026dataagentslevelsstate}
& \checkmark & $\circ$ & \checkmark & $\circ$ & -- & $\circ$ \\

\midrule
\textbf{Ours}
& \checkmark & \checkmark & \checkmark & \checkmark & \checkmark & \checkmark \\
\bottomrule
\end{tabular}
}

\end{table*}

Another body of work organizes the literature around analytical tasks,
capabilities, data modalities, or stages of the data science lifecycle
~\cite{tang2025llmagentasdataanalystsurvey,
rahman2025llmbaseddatascienceagents,
zhou2026llmscleanmesssurvey,
chang2026surveyreasoningagenticsystems}. These surveys provide important views
of what Data Agents can accomplish and, in some cases, how agent systems are
distributed across the data science lifecycle. In particular, lifecycle-based
taxonomies capture the progression from data preparation and analysis to
modeling and interpretation. Their stages, however, primarily describe the
analytical objectives being accomplished, rather than the recurring control
functions through which an agent governs its evolving task trajectory. As a
result, similar mechanisms for planning, verification, or repair may appear in
different analytical tasks without being directly compared. Related surveys also examine Data Agents from the perspective of system
architecture and increasing levels of agent capability
~\cite{luo2026dataagentslevelsstate}. Such views clarify how components and
agent architectures are assembled, but provide a different level of abstraction
from the operational mechanisms that repeatedly control individual analytical
steps.

This survey takes a complementary workflow-harness perspective. Rather than
organizing systems primarily by task, data modality, lifecycle phase, autonomy
level, or architecture, we characterize the recurring control functions through
which a Data Agent manages its task trajectory: perception, planning, execution,
verification, and repair. Within each stage, we further organize existing
methods by their technical routes, enabling mechanisms developed for different
data environments and analytical tasks to be compared at the same functional
stage. This stage-wise view also supports a process-level analysis of
reliability, allowing failures to be examined in terms of how uncertainty,
ambiguity, experience, and verification signals are propagated and handled.

\section{Conclusion}

This survey examines how Data Agents can reliably execute data science tasks through a disciplined workflow harness. We organize the literature around a five-stage loop, which consists of perception, planning, execution, verification, and repair, and identify 15 technical routes that characterize how existing systems implement these stages. Together, the workflow harness and technical route map provide a unified view of what Data Agents do and how these operations are implemented across heterogeneous data environments. Our analysis further identifies four open reliability problems: inactive semantic calibration, missing clarification, missing experience transfer, and the missing verification-repair repository. These problems highlight that reliable Data Agents depend not only on stronger models or additional tools, but also on how uncertainty, ambiguity, experience, and failure information are handled throughout the workflow.

Complementing the workflow-harness perspective, we summarize the horizontal task families supported by Data Agents, examine how these capabilities are composed across vertical application settings, and review the benchmarks used to evaluate both analytical capabilities and workflow reliability. Together, these perspectives provide a systematic view of how Data Agents operate, what capabilities they perform, how these capabilities are composed in practice, and how their reliability is evaluated, pointing toward the development of more reliable AI Data Scientists.

\bibliographystyle{ACM-Reference-Format}
\bibliography{all}
\end{document}